\documentclass{article}

\usepackage{arxiv}
\usepackage{amsmath}

\usepackage[utf8]{inputenc} 
\usepackage[T1]{fontenc}    
\usepackage{hyperref}       
\usepackage{url}            
\usepackage{booktabs}       
\usepackage{amsfonts}       
\usepackage{nicefrac}       
\usepackage{microtype}      
\usepackage{cleveref}       
\usepackage{lipsum}         
\usepackage{graphicx}
\usepackage[numbers]{natbib}
\usepackage{doi}

\title{AKF-SR: Adaptive Kalman Filtering-based  Successor Representation}

\date{}

\author{ \hspace{1mm}Parvin Malekzadeh\thanks{ Published in \textit{Neurocomputing} 467 (2022), pp.476-490. \newline \url{https://doi.org/10.1016/j.neucom.2021.10.008} } \\
	  University of Toronto\\
	\texttt{p.malekzadeh@mail.utoronto.ca} \\
	\And
	\hspace{1mm} Mohammad Salimibeni \\
	Concordia University\\
	\And
	\hspace{1mm} Ming Hou \\
	Defence Research and Development Canada \\ Toronto Research Centre
	\And
	\hspace{1mm} Arash Mohammadi \\
	Concordia University\\
	\And
	\hspace{1mm} Konstantinos N. Plataniotis \\
	University of Toronto
}

\usepackage[labelsep=endash]{caption}
\usepackage{multirow}
\usepackage{amsthm}
\usepackage{amssymb }
\usepackage{multirow}
\usepackage{tabularx}
\usepackage{bm}
\usepackage{float}
\usepackage{amsfonts}
\usepackage{flushend}
\usepackage{subcaption}
\usepackage{graphicx}
\usepackage{geometry}
\usepackage[dvipsnames]{xcolor}
\usepackage{algorithm, algpseudocode}
\usepackage{bbm}
\def\mS{\mathcal{S}}
\def\mA{\mathcal{A}}
\def\SR{\text{AKF-SR}}
\def\mP{\mathcal{P}}

\def\ua{_{a}}
\def\k{_{k}}
\def\nk{_{k+1}}
\def\i{^{i}}
\def\bt{\bm{\theta}}
\def\h{\bm{h}}

\def\K{\bm{K}}
\def\U{\bm{U}}

\def\A{\bm{A}}
\def\M{\bm{M}}
\def\I{\bm{I}}
\def\P{\bm{P}}

\def\F{\bm{F}}

\def\g{\bm{g}}
\def\X{\bm{X}}
\def\Y{\bm{Y}}

\def\W{\bm{W}}
\def\C{\bm{C}}
\def\w{\bm{w}}

\def\u{\bm{u}}

\def\um{\bm{\mu}}
\def\m{\bm{m}}
\def\Sig{\bm{\Sigma}}

\def\s{\bm{s}}
\def\pk{_{k-1}}
\def\kpk{_{k|k-1}}

\def\aL{\mathcal{L}}

\def\x{\bm{x}}

\makeatletter
\newcommand{\multiline}[1]{%
  \begin{tabularx}{\dimexpr\linewidth-\ALG@thistlm}[t]{@{}X@{}}
    #1
  \end{tabularx}
}
\begin{document}
\maketitle

\begin{abstract}
To understand animals' behavior in finding relations between similar tasks and adapting themselves to changes in the tasks, it is necessary to know how the brain generalizes the learned knowledge from a previous task to unseen tasks. Recent studies in neuroscience suggest that Successor Representation (SR)-based models provide adaptation to changes in the goal locations or reward function faster than model-free algorithms, together with lower computational cost compared to that of model-based algorithms. However, it is not known how such representation might help animals to manage uncertainty in their decision making. Existing methods for the SR learning based on standard temporal difference methods (e.g., deep neural network-based algorithms) do not capture uncertainty about the estimated SR. In order to address this issue, the paper presents a Kalman filter-based SR framework, referred to as Adaptive Kalman Filtering-based Successor Representation (AKF-SR). First, Kalman temporal difference approach, which is a combination of Kalman filter and the temporal difference method, is used within the AKF-SR framework to cast the SR learning procedure into a filtering problem to benefit from uncertainty estimation of the SR, and also decreases in memory requirement and sensitivity to model's parameters in comparison to deep neural network-based algorithms. An adaptive Kalman filtering approach is then applied within the proposed AKF-SR framework in order to tune the measurement noise covariance and measurement mapping function of Kalman filter as the most important parameters affecting the filter's performance. Moreover, an active learning method that exploits the estimated uncertainty of the SR to form the behaviour policy leading to more visits to less certain values is proposed to improve the overall performance of an agent in terms of received rewards while interacting with its environment. Experimental results based on three reinforcement learning environments illustrate the efficacy of the proposed AKF-SR framework over state-of-the-art frameworks in terms of cumulative reward, reliability, time and computational cost, and speed of convergence to changes in the reward function.
\end{abstract}

\section{Introduction} \label{sec:Introduction}
Human and animals have the learning capabilities for evaluating consequences of their actions; therefore, can adapt their behavior based on the received reward after each action~\cite{Spano,Chen, Seo}.
%
Such an adaptive behavior can be achieved in the context of Artificial Intelligence (AI) via an optimal control policy, which can be obtained by Reinforcement Learning (RL) algorithms. Generally speaking, RL is a class of Machine Learning (ML) algorithms enabling an autonomous agent to learn a task to maximize expected future discounted rewards while interacting with its environment~\cite{Parvin:access, DrMing1, DrMing2}. Unlike supervised learning, RL approaches do not need labeled data, which is generally difficult to acquire, for performing their learning processes. Traditionally, RL algorithms are categorized into two main classes: (i) Model-Free (MF) methods~\cite{Parvin:access,Venkat,Hu, kova}, which learn the value function using sample trajectories, and; (ii) Model-Based (MB) methods~\cite{2,3,Ayoub} that estimate transition and reward functions through search trees or dynamic programming. Algorithms belonging to the former category (MF algorithms) are particularly slow to adjust an agent to changes in its task (e.g., changes in the goal locations or reward function). The MB algorithms, on the other hand, can quickly adjust the agent to the changes. Such a rapid adaptation, however, comes with a high computational cost~\cite{Vertes,Sam}.  Reusing previous knowledge to facilitate learning of new tasks has been proposed as an appealing solution to the mentioned adaptation problems of MB and MF algorithms~\cite{Chen}. To improve speed of an agent for learning a new task, which results from changes in its learned tasks, without considerable increase in computational cost, the agent should be able to identify the changes and reuse its knowledge from previous tasks. Reusing the transferred information learned from the previous tasks would not be feasible if the tasks are completely irrelevant. Therefore, in this paper, we focus on the changes in the reward function while the environment remains same.

Successor Representation (SR) methods~\cite{Dayan, Ducarouge} are proposed as a potential solution to aforementioned adaptation problems of both MF and MB categories of RL algorithms in the case of changes in the reward function. The SR-based algorithms are faster to adapt to changes than MF algorithm, and provide more efficient computation than MB algorithms. The SR-based algorithms learn the expected immediate reward received after each action, together with the expected discounted future state occupancy (referred to as the SR)~\cite{Chen,Sam,Dayan}. The value function is then factorized into the SR and the immediate reward in each of successor states. This factorization enables rapid policy evaluation under changes in the reward conditions as only the reward function needs to be relearned in new tasks.

When the number of states are limited, the SR and the reward function (consequently, the value function) can be computed for each state. However, computation of the value function is not possible in RL problems with large number of states or when states are continuous; therefore, value function needs to be approximated. The function approximators in RL problems can be generally categorized into linear and non-linear function approximators~\cite{KulKarni,6}. In both cases, value of the approximate function is defined by a set of tunable parameters.
Non-linear function approximators, such as artificial neural network~\cite{KulKarni,Tang, Kim, Xie} algorithms suffer from different issues including sensitivity of the model's performance to a large number of parameters, lack of theoretical convergence guarantees, and also the need for a large number of episodes to achieve acceptable results. The linear function approximators transform the approximation problem into the problem of computing weights to fuse several local estimators. Since linear function approximators are simpler and better understood than the non-linear ones; therefore, several convergence guarantees have been provided~\cite{Dayan, Linear_1,Linear_2}. In this regard, Cerebellar Model Articulation Controllers (CMACs)~\cite{14} and Radial Basis Functions (RBFs)~\cite{13} are the commonly used linear estimators. It has been discovered that the RBFs can better represent gradual-continuous transitions in the functions to be approximated~\cite{17,18,Babu,RBF_adapt}. Due to this advantage, in this paper, we use RBFs for the SR and reward function linear approximation task.

Gershman et al.~\cite{Gershman} have established that the temporal context model, a model of episodic memory, is actually direct estimation of the SR via Temporal Difference (TD) learning algorithms. Kulkarni et al.~\cite{KulKarni} proposed a deep learning method based on the combination of TD learning with Deep Neural Networks (DNNs) for estimation of the SR. Some other works~\cite{Chen, Sam, Momennejad2017, Russek2017} focused on the development of TD learning method by using different DNNs algorithms to represent the SR. CTDL framework proposed in~\cite{Sam} combines a DNN with a Self-Organising Map (SOM) to calculate action values to improve the performance and robustness of a DNN-based RL agent. Ma et al.~\cite{Chen} proposed a DNN framework, referred to as Universal Successor Representation (USR), to approximate the SR and incorporate it with actor-critic method to learn the SR. The methods proposed in ~\cite{Chen, Sam, Momennejad2017, Russek2017}, which are based on the combination of the standard TD learning and DNNs, however, do not consider uncertainty of the SR (consequently, uncertainty in the value function), which exists at the heart of RL problems in the real world. When there is uncertainty about the environment, the agent should not be overconfident of its knowledge and exploit it all the time, but instead explore other available actions, which might be better and reduce uncertainty. The optimal solution for the exploration/exploitation trade-off is computationally intractable, but it has been shown that uncertainty can cause exploration through two different ways: by adding randomness to the value function or directing actions toward uncertain ones~\cite{32}. The estimated uncertainty of the value function, therefore, is known as a useful information for quandary between exploration and exploitation dilemma~\cite{32,23}. Lehnert et al.,~\cite{Lehnert} proposed a different method from the TD-based SR learning algorithms, where the transition probability of the RL problem is first learned; the SR is then approximated based on the learned transition model. Although the proposed framework is more sample efficient than the TD-based frameworks, the trajectory for finding the transition model is far more complex due to the difficulty in learning the accurate transition model, which makes the TD-based methods more favorable in the literature. Machado et al.~\cite{count} has recently proposed a framework, known as the Substochastic Successor Representation (SSR), which uses the norm of the SR as an exploration bonus. In the proposed scheme, first, the SSR is learned by minimizing the TD error through a DNN. The value function is then learned through a DNN structured similar to DQN framework~\cite{7}, which adopts the estimated SSR for exploration/exploitation trade-off. Geist et al.~\cite{23} proposed Kalman Temporal Difference (KTD) framework for uncertainty estimation of the value function, but in the context of MF methods. The proposed KTD framework also benefits from less sensitivity of the framework's performance to model's parameters and reduced time and memory requirement for finding and learning of the best model in comparison to DNN-based frameworks~\cite{Chen, Sam, Momennejad2017, Russek2017, count}. Less sensitivity to parameter settings improves the reproducibility aspect of a reliable algorithm to regenerate more consistent performances across multiple learning runs; therefore, reduces the risk of generating unpredictable performances in different practical applications~\cite{Chan}. Geerts et al.~\cite{Geerts} and Salimibeni et al.~\cite{Mohammad:Icassp} applied KTD framework for the SR estimation in RL problems, respectively, with discrete state spaces and multiple agents. The proposed algorithms, however, do not use the uncertainty information of the estimated SR, which can be achieved from KTD algorithm, for the action selection process. Since value function is computed as a dot product between the SR and the reward function, there is a need for approximation of the reward function. However, Geerts et al.~\cite{Geerts} do not discuss the reward function learning process. In this paper, as an initial step, we adopt the KTD framework for the SR learning process and propose a reward learning algorithm. Then, an innovative action selection scheme benefiting from the achieved uncertainty/belief of the estimated SR in order to deal with exploration/exploitation dilemma.

As stated previously, within the SR domain, for computation of the value function, both the SR and reward function need to be learned. So far, we proposed application of KTD within this context for the SR learning. To learn the reward function, on the other hand, we develop a Kalman Filter (KF) to estimate the RBFs' weights. The challenge here is selection of KF's parameters. Performance of KF-based algorithms are highly dependent on the filter's parameters. It has been shown that measurement noise covariances of a KF is one of the most important parameters of the filter and its improper selection can significantly degrade the filter's act and even cause divergence of the filter~\cite{27,28}. In the earlier studies, the underlying parameter assumed to be constant during the estimation process with its value being adapted manually by trial and error. Such a bruteforce approach, however, is significantly challenging due to variation of the parameter value. Lately, multiple studies have proposed to adjust the parameter value at each step via Adaptive Kalman Filter (AKF) approaches to enhance the overall accuracy of the filter~\cite{Akhlaghi}. Generally speaking, AKF approaches can be categorized into two main groups: (i) Innovation-based Adaptive Estimation (IAE) methods~\cite{24}, and; (ii) Multiple Model Adaptive Estimation (MMAE) methods~\cite{25}. The former category uses a single Extended Kalman Filter (EKF) to adapt measurement noise covariances based on the innovation or the residual sequence. In the IAE methods, perfect knowledge of the system's model and an appropriate window size is required for the parameter computation. The MMAE methodologies, on the other hand, use a weighted sum of multiple KFs running in parallel for the parameter adaptation addressing the aforementioned issue. Consequently, the MMAE methods have attracted more attention owing to their capabilities to manage parametric uncertainty and their independence from imposing specific requirements on the system model. Capitalizing on the success of MMAE frameworks, in this paper, we develop an innovative MMAE-based modeling framework for learning of the reward function within the SR context. Different algorithms were proposed for fusion of the KFs in MMAE systems~\cite{26}, among which the classical scheme is used in this paper due to its exponentially fast convergence speed to the best candidate model (filter). In the classical MMAE scheme, weight of each filter is achieved from a Bayesian approach.

Duration of the learning process and quality of the learned policies, which are obtained based on the proposed combination of KTD and MMAE techniques, highly depend on managing a trade-off between exploration and exploitation. Too much exploration prevents from maximizing the immediate rewards since the selected actions may result in a negative reward from the environment. On the other hand, exploiting uncertain knowledge offered by the environment reduces the expected future rewards because actions are selected given current information; therefore, may not be the optimal ones~\cite{Michel}. As the value function is modeled as a function of stochastic variables; therefore, there is a stochastic variable for each state-action pair. The dilemma between exploration and exploitation should profit from such uncertain information~\cite{23}. In this paper, we propose an active learning scheme, which uses the KTD framework to tackle the uncertainty computation, an important issue which has been overlooked in the literature.

In summary, this paper proposes a SR-based framework, referred to as Adaptive Kalman Filtering-based Successor Representations ($\SR$), which can adapt quickly to changes in the reward function or goal locations faster than MF methods and with the lower computational cost compared to MB algorithms. The following $4$ key contributions are made in this paper:
\begin{itemize}
\item RBFs estimators are incorporated within the $\SR$ framework to project continuous states into feature vectors such that the SR and the reward function can be modeled as linear functions of the feature vectors.
\item Within the proposed $\SR$ framework, we adopt MMAE and gradient descent-based schemes for the reward function estimation (learning) via KF, which respectively compensate for the insufficient information about the measurement noise covariance and measurement mapping function of the KF as the most important parameters of a KF.
\item The KTD framework is used within the $\SR$ framework, which casts the SR learning procedure into a filtering problem to estimate uncertainty of the learned SR. Moreover, by applying KTD, we benefit from decreases in memory and time spent for the SR learning and also sensitivity of the framework's performance to its parameters (i.e., more reliable)  when compared with DNN-based algorithms.
\item An active learning scheme is exploited to balance the amount of exploration and exploitation based on uncertainty information of the value function obtained from KTD algorithm of the SR learning. The proposed active learning mechanism effectively improves performance of the proposed $\SR$ framework in terms of cumulative reward as shown via three RL platforms.
\end{itemize}
The remainder of the paper is organized as follows: In Section~\ref{sec:PrbFor}, an overview of RL and SR is presented. The proposed $\SR$ framework is developed in  Section~\ref{sec:AKF-SR}. Section~\ref{sec:Sim} presents experimental results obtained based on three benchmark RL platforms illustrating effectiveness  of the proposed $\SR$ framework. Finally, Section~\ref{sec:con} provides conclusion of the paper.

\section{Problem Formulation} \label{sec:PrbFor}
In this section, first, the required background to follow developments presented in the remainder of the paper will be presented. In what follows, we use following notation: Scalar variables are represented by Non-bold letter (e.g., $X$ or $x$)); Vectors are represented by lowercase bold letter (e.g., $\x$); Matrices are denoted by capital bold letter (e.g., $\X$), and transpose of matrix $\X$ is shown by $\X^T$.

\subsection{Reinforcement Learning (RL)}
Generally speaking, the main objective of an autonomous agent in RL models is to learn an optimum control policy through interaction with the dynamic system (also considered as the agent's environment)~\cite{32}. More specifically, within a RL context, an autonomous agent selects actions from the action set $\mA$ by following an optimal policy $\pi^*$ in such a way that its cumulative reward will be maximized over time of the agent's interaction with its environment. At time step $k$, the autonomous agent takes an action $a\k \in \mA$ given its current state $\s\k \in \mS$ based on a given policy $\pi(.|\s\k)$, which maps the state $\s\k$ to a probability distribution over the action space $\mA$. If the policy is
deterministic, the agent will always map to one specified action in a given state. The environment then answers to the selected action by taking the agent to state $\s\nk \in\mS$ with the transition probability of $\text{Pr}(\s\nk| \s\k, a\k)$ and returning a reward $R(\s\k,a\k)$ to the agent, where $R(.)$ is the reward function. Given initial state $\s_0=\s$ and action $a_0=a$, $R(\s_0,a_0)=R(\s,a)$ has no randomness; however, $R(\s\k,a\k)$ for $k \geq 1$ is a function of random variable $\s\k \sim \text{Pr}(.| \s\pk, a\pk)$ and possibly random variable $a\k \sim \pi(.|\s\k)$ (i.e., possibly stochastic policy). The expected value of the reward function $R(\s\k,a\k)$ for ($k \geq 1$) is, therefore, calculated by taking the expectation with respect to the probabilities $\text{Pr}(.| \s\pk, a\pk)$ and $\pi(.|\s\k)$.
The rewards are discounted using discount factor $\gamma\in (0,1)$ to control the importance of the future rewards versus the immediate ones. The $5$-tuple $\{\mS, \mA, \mP\ua, R, \gamma\}$ defines a Markov Decision Process (MDP), which provides a rigorous mathematical framework for modeling the RL tasks. For a fixed policy $\pi$, the return $\sum_{k=0}^{\infty}\gamma^k R(\s\k,a\k)$, represents the sum of discounted rewards observed along one trajectory of states while following $\pi$. The state-action value function $Q^{\pi}(\s,a)$ for a policy $\pi$ estimates expected value of the return $\sum_{k=0}^{\infty}\gamma^k R(\s\k,a\k)$ over all infinite length trajectories that start at state $\s_0=\s$ with action $a_0=a$, then acting according to $\pi$:
\begin{eqnarray}
 \lefteqn{\!\!\!\!\!\!\!\!\!\!\!\!\! Q^{\pi}(\s, a) =  \mathbb{E}\left\{\sum_{k=0}^{\infty}\gamma^k R(\s\k,a\k) \right\}\label{Eq:Q}} \\
&&\!\!\!\!\!\!\!\!\!\!\!\!\!\!\!\!\!\!\!\!\!\! \s\k \sim \text{Pr}(.|\s\pk,a\pk), a\k \sim \pi(.|\s\k), a_0=a , \s_0=\s. \nonumber
\end{eqnarray}
Using the linearity of the expectation, the expectation function $\mathbb{E}$ in Eq.~\eqref{Eq:Q} is, therefore, calculated over $\text{Pr}(.| \s\pk, a\pk)$ and possibly stochastic policy $\pi(.|\s\k)$ for all $k \geq 1$, i.e.,
\begin{eqnarray}
  \!\!\!\!\!\!\!\!\!\!\!\!\!\!\!\! Q^{\pi}(\s, a) &=& \mathbb{E}\left\{\sum_{k=0}^{\infty}\gamma^k R(\s\k,a\k) \right\} = \sum_{k=0}^{\infty} \gamma^k \mathbb{E} \left\{R(\s\k,a\k)\right\}\,\,\,\,\,\,\,
\\
\!\!\!\!\!\!\!\!\!\!\!\!\!\!\!\! &=& R(\s,a) + \sum_{k=1}^{\infty} \gamma^k \mathbb{E} \left\{R(\s\k,a\k)\right\}, \nonumber
\end{eqnarray}
where $\s\k \sim \text{Pr}(.|\s\pk,a\pk)$,  $a\k \sim \pi(.|\s\k), a_0=a $, and $\s_0=\s$.
The state-action value function ($Q^{\pi}(\s,a)$) can be gradually update based on the Bellman update idea~\cite{31}. The iterative update of the state-action value function forms a learning algorithm known as Temporal Difference (TD) learning~\cite{32,Xia}, i.e.,
\begin{eqnarray}
\lefteqn{\!\!\!\!\!\!\!\!\!\!Q^{\pi}_{\text{new}}(\s\k, a\k) = Q^{\pi}_{\text{old}}(\s\k, a\k) }\nonumber \\
&&\!\!\!\!\!\!\!\!\!\!\!\!\!\!\!\!\!\!+ \alpha \Big(R(\s\k,a\k) +\gamma\,Q^{\pi}_{\text{old}}(\s\nk, a\nk)  - Q^{\pi}_{\text{old}}(\s\k, a\k) \Big),
\label{Eq:TD}
\end{eqnarray}
where $Q^{\pi}_{\text{old}}(\s\k, a\k)$ and $Q^{\pi}_{\text{old}}(\s\nk, a\nk)$ represent the latest estimates of the state-action value function at state-action pairs ($\s\k,a\k$) and ($\s\nk,a\nk$), respectively. $Q^{\pi}_{\text{new}}(\s\k, a\k)$ is an updated representation given an observed transition ($\s\k,a\k,\s\nk, R(\s\k,a\k)$) and ($R(\s\k,a\k) +\gamma\,Q^{\pi}_{\text{old}}(\s\nk, a\nk)  - Q^{\pi}_{\text{old}}(\s\k, a\k)$) is known as the TD error that intuitively represents the difference between the predicted reward
according to the current estimate of the state-action value function and the actual observed reward at time step $k$. The parameter ($0<\alpha\leq 1$) is the learning rate determining how much of the error should we accept to adjust our estimates towards. During the learning process of state-action value function, actions are selected using the current policy $\pi$. After convergence of the learning process (i.e., the TD error converges to zero) where optimal policy $\pi^*$ is achieved, actions are obtained based on the optimal policy $\pi^*$ as follows
\begin{eqnarray}
a = \arg\max_{a\in\mA}Q^{\pi}(\s\k, a).\label{Eq:action}
\end{eqnarray}
This completes a brief introduction to basics of RL approaches. Next, we present the SR modeling framework.

\subsection{The Successor Representation (SR)}
Dayan~\cite{Dayan} claimed that distances between states can express similarities of future paths of an agent given the policy $\pi$. Successor Representation (also referred to as SR)~\cite{Dayan} estimates the cumulative time expected to be spent in future state $\s'$ given the initial state $\s$ and initial action $a$ by following policy $\pi$:
\begin{eqnarray}
\M^{\pi}(\s,\s',a) =\mathbb{E}
\left\{\sum_{k=0}^{\infty}\gamma^k \mathbbm{1}[\s\k=\s']|\s_0=\s, a_0=a\right\},\label{Eq:7}
\end{eqnarray}
 In a discrete state-space scenario, the SR ($\M^{\pi}(:,:,a)$) is a ($|\mS|\times |\mS|$) matrix, where $|\mS|$ is cardinality of $\mS$. Gershman et al.~\cite{Gershman} showed that similar to the TD learning of state-action value function in Eq.~\eqref{Eq:TD}, the SR can be also updated in an on-policy recursive form~\cite{Gershman}, known as SARSA, as follows
\begin{eqnarray}
\!\!\M^{\pi}_{\text{new}}(\s\k,\s',a\k) = \M^{\pi}_{\text{old}}(\s\k,\s',a\k) ~+  
\alpha \Big(\mathbbm{1}[\s\k=\s'] +\gamma\,\M^{\pi}_{\text{old}}(\s\nk,\s',a\nk) - \M^{\pi}_{\text{old}}(\s\k,\s',a\k) \Big).\label{Eq:TD_SR}
\end{eqnarray}
where $\M^{\pi}_{\text{old}}(\s\k,\s', a\k)$ and $\M^{\pi}_{\text{old}}(\s\nk,\s',a\nk)$  represent the latest estimates of the expected number of times that the agent visit state $\s'$ given initial state-action pair ($\s\k,a\k$) and ($\s\nk,a\nk$), respectively. The TD error here represents the error in state
visitation count of state $\s'$ starting at initial state-action pair ($\s\k,a\k$).
Given the SR, the state-action value function, Eq.~\eqref{Eq:Q}, can be expressed as the inner product of the SR and the estimated reward function~\cite{Dayan}, i.e.,
\begin{eqnarray}
Q^{\pi}(\s\k, a\k) = \sum_{\s'\in \mS} \sum_{a'\in \mA}  \M^{\pi}(\s\k,\s',a\k) R(\s',a').
\label{Eq:Q_SR}
\end{eqnarray}
Eq.~\eqref{Eq:Q_SR} demonstrates the most important feature of SR-based frameworks, as it represents a linear mapping that allows the state-action value function $Q^{\pi}(\s\k, a\k)$ to be reconstructed straightforwardly based on reevaluation of the reward function ($R(\s',a')$) in case of changes in the reward function. Next, we focus on extending SR model (Eqs.~\eqref{Eq:7}-\eqref{Eq:Q_SR}) to scenarios with continuous state spaces.

\subsection{Feature-based Successor Representations}\label{sec:kalman}
Estimation of the SR and the reward function are impractical for each state in RL problems with large number of states or continuous state spaces; hence, as such one needs to resort to their approximated forms. In such cases, each pair $(\s,a)$, is mapped into a $L$-dimensional state-action feature vector $\bm{\psi}(\s,a)$ ($\bm{\psi}: \mA \times \mS \rightarrow \mathbb{R}^{L}$).
In this setting, the successor representation is generalized to a feature-based SR vector~\cite{Barreto}, which encodes the expected feature values as follows
\begin{eqnarray}
\m^{\pi}(\s,a)=\mathbb{E}
\left\{\sum_{k=0}^{\infty}\gamma^k \bm{\psi}(\s\k,a\k)|\s_0=\s, a_0=a\right\}, \label{Eq:Sr_appr}
\end{eqnarray}
For approximation of the SR and reward function, linear function approximators are reasonable choices due to their simpler implementations and faster computation speed compared to the non-linear ones. The convergence of linear function approximators have been already guaranteed in RL problems~\cite{Dayan, Linear_1,Linear_2}. A linear function of the feature vectors is, therefore, used to approximately factorize the immediate reward function for the pair $(\s,a)$ as follows
\begin{eqnarray}
R(\s\k,a\k)\approx\bm{\psi}(\s\k,a\k)^T \bt\k, \label{Eq:reward}
\end{eqnarray}
where $\bt\k$ is the reward's weight vector. The state-action value function (Eq.~\eqref{Eq:Q_SR}), therefore, can be computed as
\begin{eqnarray}
Q^{\pi}(\s\k,a\k)\approx \bt^T\k \m^{\pi}(\s\k,a\k). \label{Eq:Q_estimation}
\end{eqnarray}
We assume that the SR vector $\m^{\pi}(\s\k,a\k)$ in Eq.~\eqref{Eq:Sr_appr} can be also approximated as a linear function of the same feature vector:
\begin{eqnarray}
\m^{\pi}(\s\k,a\k)\approx \W\k\, \bm{\psi}(\s\k,a\k), \label{Eq:linear_SR}
\end{eqnarray}
where $\W\k$ is a ($L\times L$) matrix embodied by the weights $W_{i,j}$. The TD learning of the SR in Eq.~\eqref{Eq:TD_SR} is update as
\begin{eqnarray}
\m^{\pi}_{\text{new}}(\s\k,a\k)=\m^{\pi}_{\text{old}}(\s\k,a\k)+\alpha\big(\bm{\phi}(\s\k,a\k)+\gamma \m^{\pi}_{\text{old}}(\s\nk,a\nk)-\m^{\pi}_{\text{old}}(\s\k,a\k)\big).
\label{Eq:TD_SR2}
\end{eqnarray}
Our discussion on background of the SR is complete. Next, the proposed $\SR$ framework is presented. In the remainder of this paper, for simplicity, we assume that the policy and transition function of the system are deterministic.
\section{AK-SR: Adaptive Kalman Filtering-based Successor Representations} \label{sec:AKF-SR}
%
Once the approximation structure of the SR and the reward function have been defined, a suitable algorithm needs to be designed to learn (estimate) the reward function $R(\s,a)$ and the SR vector $\m^{\pi}(\s,a)$. For these estimations, sample transition of the system, the feature vectors, and the received reward from the environment are used as the measurements.
DNN-based RL methods~\cite{ Chen,KulKarni, count, 7, Chan}, require to store all these measurements together with the network's parameters and activations to be learned in batches. In the learning process, one needs to store the activations from a forward propagation to be utilized later for computation of the error gradients in a back propagation process. Therefore, considerably high memory is required for implementation of deep learning-based techniques. For instance, there are $26$ million parameters in a $50$-layer ResNet network resulting in the need to compute $16$ million activations during the forward pass. Approximately, the memory required for training of a ResNet-$50$ network with a mini-batch of $32$, is over $7.5$ GB of local DRAM. Furthermore, similar to other standard TD learning-based algorithm, most of the DNN-based frameworks~\cite{Sam, KulKarni, 7} do not consider uncertainty within the value approximation context. The difficulty between exploration and exploitation should use from such uncertainty information. Finally, reliability of a learning algorithm is another important factor which needs to be verified for applications to real word scenarios. For a learning process to be reliable in different applications, it should be capable of regenerating consistent performances over multiple runs with the least frangibility to the model's parameters (reproducibility aspect of reliability)~\cite{Chan}. However, performance of a DNN model, is highly affected by its large number of parameters required to be tuned. Tuning of such a large number of parameters, therefore, leads to high sensitivity and considerable time and effort needed to tune the parameters, which make DNN-based RL algorithms unreliable for practical applications.

By contrast, filtering algorithms~\cite{AK1,AK2}, which are efficient techniques to process sequential data, can be implemented based only on the last measurement. Such approaches eliminate the necessity of the learning process to record the complete measurement history, which, in turn, translates into significant reduction in time and memory requirements in comparison to DNN-based techniques. It also has been shown that applied filtering-based algorithms (such as KTD~\cite{23}) estimate uncertainty of the value function and consider that for action selection during the learning process in order to make a balance between exploring unknowns and exploiting the agent's knowledge. Furthermore, despite DNNs, a small number of parameters is required to be tuned in filtering-based methods resulting in less time and effort required for the parameters tuning and also less vulnerability of its performance to the model's parameters in comparison to its DNN-based counterpart.
The performance of filtering-based algorithms, however, are highly related to the filter's parameters and complete information about theses parameters is not accessible. In order to tackle this problem and tune the filter's parameters, earlier studies proposed adaptive multiple model filters~\cite{AK3, AK4} to adapt the filter's parameters.

\begin{figure}
\centering
\captionsetup{justification=centering}
\includegraphics[scale=0.55]{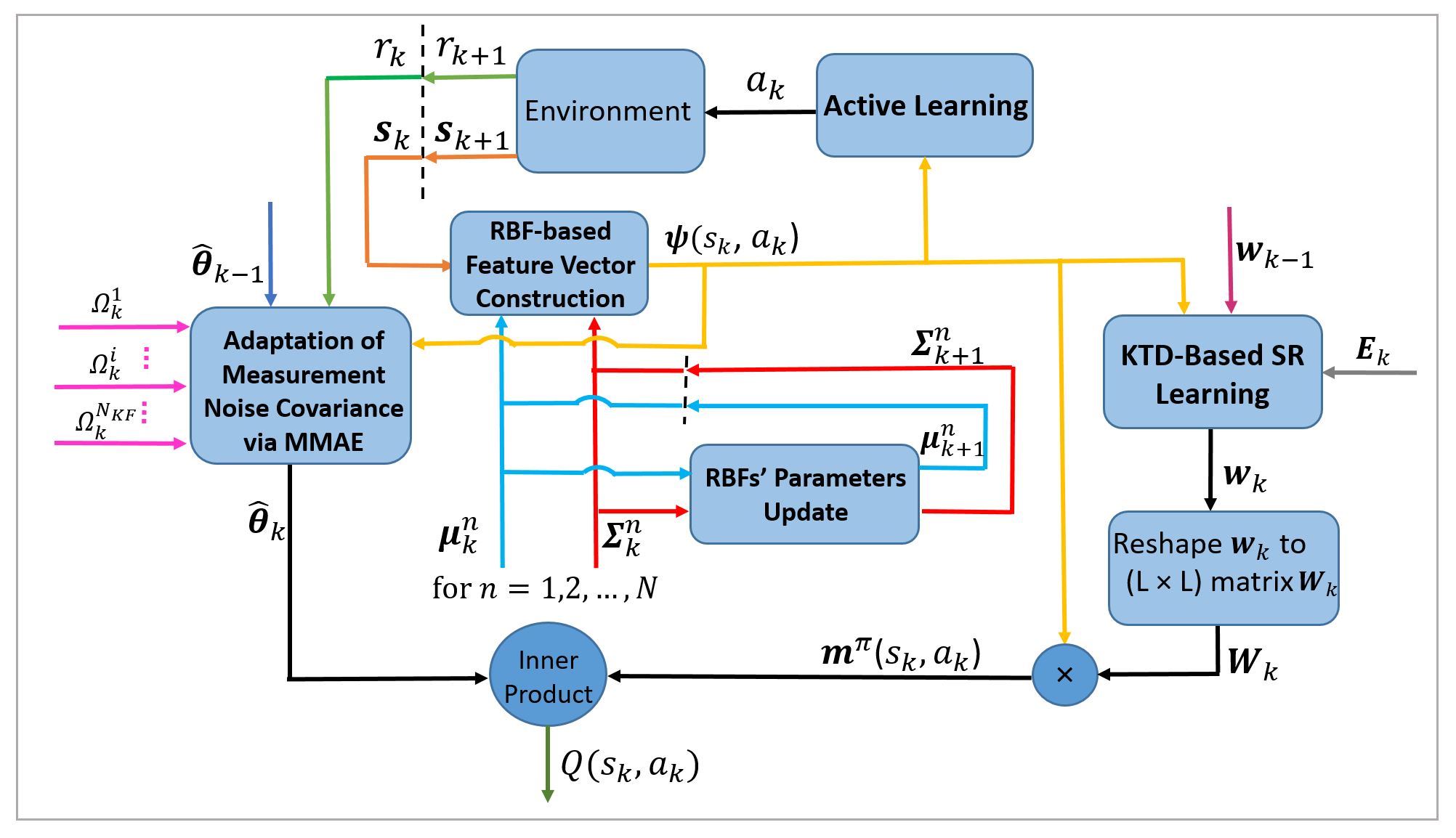}
\caption{\small Block diagram of the proposed $\SR$ framework.}\label{Fig:structure}
\end{figure}

The proposed $\SR$ framework is, therefore, based on development of KTD framework to the SR learning and adaptable KF in order to provide a powerful efficient means for estimation the reward function. The $\SR$ consists of the following four main modules as follows:
\begin{enumerate}
\item\textit{RBF-based Feature Vector Construction}, which projects a pair ($\s,a$) into a feature vector consisting of RBFs in order to generalize the SR and the reward function (consequently, value function) to continuous state spaces such that the SR and the reward function can be modeled as linear functions of the feature vectors.
\item\textit{Reward Learning}, which estimates the reward weight vector $\bt$ via a KF. To tune the measurement noise covariance of the KF a multiple model adaptive estimation method is utilized within KF formulation for reward learning.
Furthermore, restricted gradient descent is adopted which regularizes the measurement mapping function of the KF by updating the means and covariances of the underlying RBFs.
\item\textit{KTD-based SR Learning}, which estimates the SR weight matrix $\W$ using KTD algorithm in order to estimate uncertainty of the SR (consequently, the value function).
\item\textit {An Active Learning Process}, which uses the value function's uncertainty achieved from utilized KTD in the SR learning process to select the action that reduces the system's uncertainty more than any other potential actions.
\end{enumerate}
Fig.~\ref{Fig:structure} provides a block diagram of the proposed $\SR$ framework.
The aforementioned four components of the proposed $\SR$ framework are detailed in the following sub-sections.
%
\subsection{RBF-based Feature Vector Construction}
As mentioned earlier, in RL problems with large/continuous state spaces, we need to represent each pair ($\s,a$) with a state-action feature vector in order to approximate value function for all of the states in $\mS$. In this study, state $\s$ is mapped into a $N_{\text{RBF}}$-dimensional state feature vector $\bm{\phi}(\s)$ consisting of basis functions $\phi_n(\s)$ as follows
\begin{eqnarray}
\bm{\phi}(\s\k) = \big[\phi_{1}(\s\k), \phi_{2}(\s\k),\ldots , \phi_{N-1}(\s\k), \phi_{N_{\text{RBF}}}(\s\k)\big]^T, \label{Eq:28}
\end{eqnarray}
where each element of vector $\bm{\phi}(\s\k)$ in Eq.~\eqref{Eq:28} is selected as a radial basis function given by
\begin{eqnarray}
\phi_{n}(\s\k) &=& e^{\frac{-1}{2}(\s\k-\um^{n}\k)^T\big({\Sig}^{n}\k\big)^{-1}(\s\k-\um^{n}\k)},\label{Eq:phi}
\end{eqnarray}
where  the mean and covariance of the RBFs at step $k$ are represented by $\um\k^{n}$ and $\bm{\Sigma}\k^{n}$. The state-action feature vector $\bm{\psi}(\s\k,a\k=a_d)$ for ($1 \leq d \leq N_{\text{actions}}$), where $N_{\text{actions}}$ is the total number of actions, is then generated from $\bm{\phi}(\s\k)$ by assigning this state feature vector to the corresponding spot for action ($a\k=a_d$) while the feature vector values for the remainder of the actions are set to zero, i.e.,
\begin{eqnarray}
\bm{\psi}(\s\k,a\k=a_d) = [0,\ldots 0, \phi_{1}(\s\k), \ldots, \phi_{N_{\text{RBF}}}(\s\k),0,\ldots 0]^T.\label{Eq:si}
\end{eqnarray}
For each pair ($\s,a$), the generated state-action feature vector $\bm{\psi}(\s,a)$ is, therefore, a vector of size $L= N_{\text{RBF}} \times N_{\text{actions}}$.

After construction of the state-action feature vector $\bm{\psi}(\s\k,a\k)$ through Eq.~\eqref{Eq:si}, the reward weight vector $\bt\k$ in Eq.~\eqref{Eq:reward} and the SR weight matrix $\W\k$ in Eq.~\eqref{Eq:linear_SR} need to be learned. Sub-sections~\ref{Sec:reward} and~\ref{Sec:SR} detail the learning process of the reward function and the SR, respectively.
\subsection{Reward Learning}\label{Sec:reward}
As mentioned in Sub-section~\ref{sec:kalman}, the reward function can be approximated as a linear function of the state-action feature vector $\bm{\psi}(\s\k,a\k)$ with the weigh vector of $\bt\k$. In the proposed $\SR$ framework, the reward's weight vector $\bt\k$ considered as a random variable, which can be estimated via KF with the following measurement model
\begin{eqnarray}
R(\s\k,a\k) &=& \underbrace{\bm{\psi}^T(\s\k,a\k)}_{\h\k}\,\bt\k + X\k, \label{Eq:reward_update}
\end{eqnarray}
where $\h\k$ is the measurement mapping function, and $X\k$ is a zero-mean Gaussian noise with unknown variance of $\Omega\k$. The evolution of the reward function is supposed to be determined by a so-called evolution equation, which connects the current value $\bt\k$ with the previous one $\bt\pk$. However, true dynamic of the reward function cannot anyway be obtained in a general case. Following earlier works\cite{Parvin:access, Dayan, Geerts}, where the dynamic is assumed to be a random diffusion process, such that the passage of time increases uncertainty without changing the mean belief/uncertainty about the estimated $\bt$, in this paper, a heuristic evolution model following Occam razor principle is adopted as follows
\begin{eqnarray}
\bt\k &=& \F\k \bt\pk + \bm{b}\k \label{Eq:reward_predict},
\end{eqnarray}
where $\F\k = 0.9 \I_L$ in order to make the filter stable, and $\bm{b}\k$ is supposed to be a white Gaussian noise with zero mean and covariance of $\bm{B}\k$. Given the assumptions of the linear evolution model and additional noise with Gaussian distribution, a KF is guaranteed to provide the optimal solution. For solving this KF problem, after the initialization step, the weights and their associated  covariance matrices are computed as follows
\begin{eqnarray}
\hat{\bt}\kpk &=& \F\k \hat{\bt}\pk, \label{Eq:16}\\
\text{and }~~\P_{k|k-1} &=& \F\k \P_{k-1} \F\k^T+ \bm{B}\k. \label{Eq:17}
\end{eqnarray}
Then, the received reward $r\k=R(\s\k,a\k)$ from the environment is utilized to update the estimates as follows
\begin{eqnarray}
\!\!\!\!\!\!\!\!\!\!\K\k &=& \P_{k|k-1} \h\k^T \big(\h\k\P_{k|k-1}\h\k^T +\Omega\k \big)^{-1},\label{Eq:18}\\
\!\!\!\!\!\!\!\!\!\!\hat{\bt}\k &=& \hat{\bt}\kpk + \K\k\big(r\k - \h\k\hat{\bt}\kpk \big), \label{Eq:19}\\
\!\!\!\!\!\!\!\!\!\!\text{and }~~\P_{k} &=& \big(\I - \K\k\h\k\big)\P_{k|k-1}.\label{Eq:20}
\end{eqnarray}
Note that the performance of KF is highly affected by the filter's parameters and improper selection of the parameters can cause the instability of the process for applications in practical RL problems. Complete information about these parameters is, typically, not available in practical scenarios. In most of previous studies, it has been supposed that these parameters were fixed during the estimation and were manually adapted by trial and error. Due to changes of noise levels in different contexts, however, it can be difficult to utilize such a method to set up the correct value of the parameters.
Within a KF framework, the measurement noise covariance matrix and the measurement mapping function  are the most critical parameters as they control flow of new information. Improper choices of theses parameters could degrade performance of the KF, even cause divergence of the system, and are, therefore, considered for adaptation in this work. In the following, we represent adaptation process of measurement noise covariance $\Omega$. Adaptation of the mapping function $\h$ is discussed in Sub-section~\ref{Sec:RBFs}. 
\subsubsection{Adaptation of Measurement Noise Covariance via MMAE}\label{Sec:MMAE}
For adaptation of the measurement noise covariance, MMAE scheme is used via implementation of a set of KFs, where different values for $\Omega$ is considered in each mode-matched filter. Eqs.~\eqref{Eq:18}-\eqref{Eq:20} are then modified as follows
\begin{eqnarray}
\!\!\!\!\!\!\!\!\!\!\K\i\k &\!\!\!\!=\!\!\!\!& \P_{k|k-1} \h\k^T\big(\h\k\P_{ k|k-1}\h\k^T +\Omega\k\i \big)^{-1} \label{Eq:MMAE:1}\\
\!\!\!\!\!\!\!\!\!\!\hat{\bt}\i\k &\!\!\!\!=\!\!\!\!& \hat{\bt}\kpk + \K\i\k\big(r\k - \h\k \hat{\bt}\kpk\big)\label{Eq:MMAE:2}\\
\P\i_{k} &\!\!=\!\!& \big(\I - \K\i\k\h\k\big)\P_{ k|k-1},\label{Eq:MMAE:3}
\end{eqnarray}
where superscript $i$ for ($1 \leq i \leq N_{\text{KF}}$), shows the $i^{\text{th}}$ filter within the bank of $N_{\text{KF}}$ Kalman filters. The $i^{\text{th}}$ mode-matched filter, denoted by $m\i$, exploits $\Omega\k\i$ as its measurement noise covariance. The output posteriori of localized matched filters are then averaged based on their associated normalized weights $\omega\i\k$ as follows
\begin{eqnarray}
\text{Pr}(\bt\k|\Y\k) &=& \sum_{i=1}^{N_{\text{KF}}}\omega\i\k \text{Pr}(\bt\k|\Y\k, \Omega\k\i), \label{Eq:24}
\end{eqnarray}
where $\Y\k$ shows the reward sequence $\{r_{1},r_2, \ldots, r\k\}$.
The weight of mode $m^{i}$ is calculated recursively using the Bayesian rule as
\begin{eqnarray}\label{eq:modelProb3}
\!\!\!\!\!\!\!\!\!\!\!\!\!\!\!\!\!\!\!\!\!\!\omega\i\k  \triangleq \text{Pr}\big(m\i\k|\Y\k\big) &\!\!\!\!\!=\!\!\!\!&  \frac{P\big(r\k| \Y\pk, m^{i}\k\big) P\big(m^{i}\k|\Y\pk\big) }
{\sum_{j=1}^{N_{\text{KF}}} P\big(r\k| \Y\pk, m\k^{j}\big) P\big(m\k^{j}|\Y\pk\big)},
\end{eqnarray}
where the denominator is just a normalizing factor to make sure that $P(m^{i}\k|\Y\k)$ is an appropriate probability density function (PDF). Term  $\aL\i\k \triangleq \text{Pr} \big(r\k|\Y\pk, m^{i}\k\big)$ in  the nominator is the likelihood function of mode $i$, which is calculated as a PDF of Kalman filter measurement residual ($\epsilon\k=r\k-\h\k \hat{\bt}\kpk$) as follows
\begin{eqnarray} \label{Eq:25}
\lefteqn{\!\!\!\!\!\!\!\!\!\!\!\!\!\!\! \aL\i\k = \text{Pr}(r\k|\hat{\bt}\kpk,\Omega\k\i)= \frac{1}{\sqrt{\text{det}\big[2\pi Z\k\i\big]}}.e^{\frac{-1}{2}\epsilon\k^T\big(Z\k\i\big)^{-1}\epsilon\k}\nonumber,}\\
\lefteqn{\!\!\!\!\!\!\!\!\!\!\!\!\!\!\!\text{where}\quad Z\k\i = \h\k\P_{\bt,k|k-1}\h\k^T+ \Omega\k\i.}
\end{eqnarray}
Eq.~\eqref{eq:modelProb3} for computing $\omega\k\i$ is, therefore, reduced to
\begin{eqnarray}\label{eq:core2}
\omega\i\k &\!\!=\!\!& \frac{\omega\i\pk\aL\i\k}{\sum_{j=1}^{M}\omega^{j}\pk\aL^{j}\k}.
\end{eqnarray}
The initial value of the weights are set to $\omega\i_0=1/N_{\text{KF}}$ for $i=1, 2, \dots, M$. The posteriori estimate $\hat{\bt}\k$ and its error covariance are then obtained as
\begin{eqnarray}
\hat{\bt}\k &\!\!=\!\!& \sum_{i=1}^{N_{\text{KF}}}\omega\i\k\hat{\bt}\i\k,\\
\P_{k} &\!\!=\!\!& \sum_{i=1}^{N_{\text{KF}}}\omega\i\k\left(\P_{\bt,k}\i + (\hat{\bt}\i\k-\hat{\bt}\k)(\hat{\bt}\k\i-\hat{\bt}\k)^T \right).\label{eq:n25}
\end{eqnarray}
%
\begin{figure}
\centering
\captionsetup{justification=centering}
\includegraphics[scale=0.59]{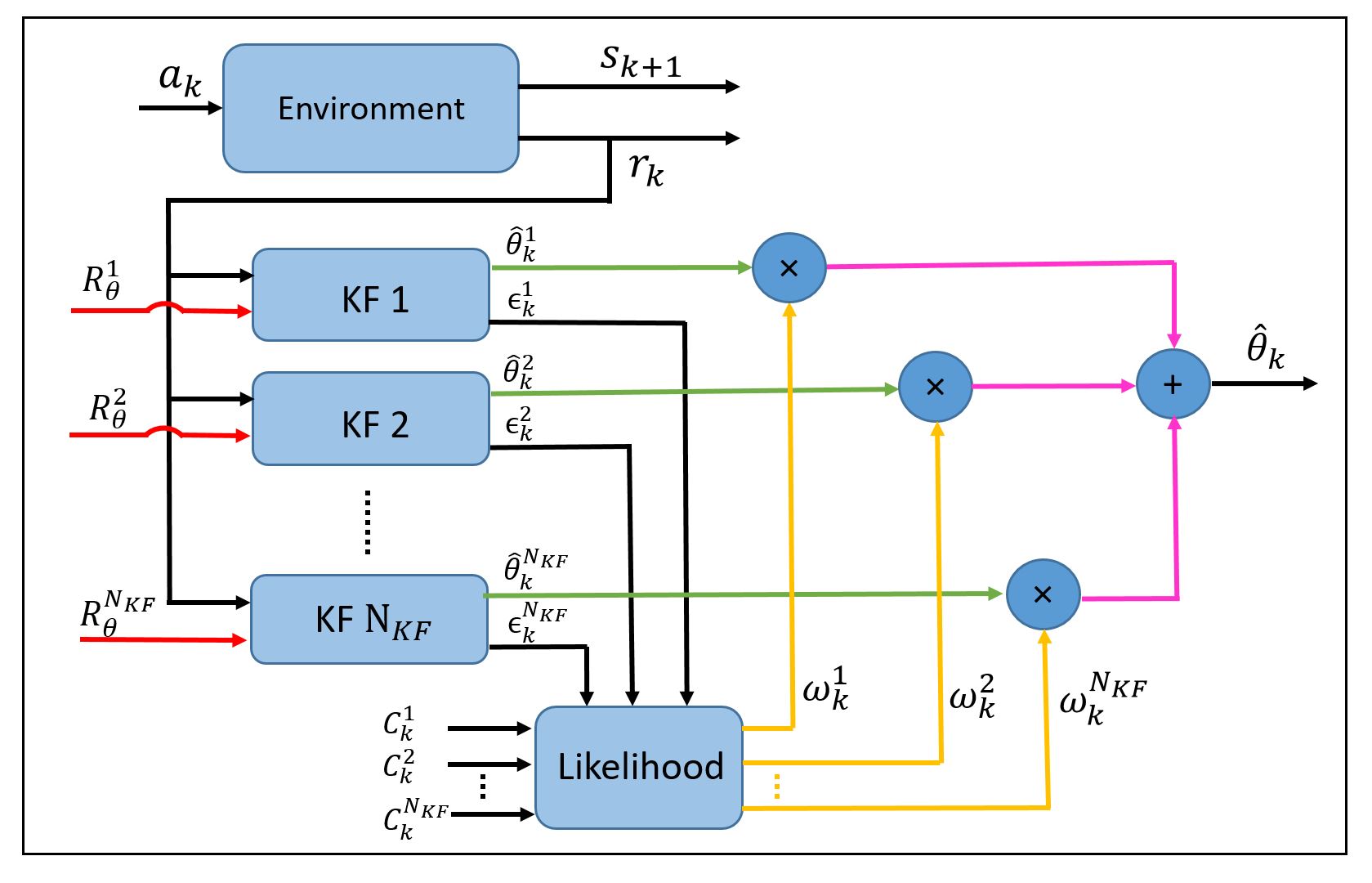}
\caption{\small MMAE structure.}\label{Fig:MMAE}
\end{figure}
The MMAE process is shown in Fig.~\ref{Fig:MMAE}. This completes presentation of the proposed MMAE-based algorithm for reward function learning. Next, we focus on adaptation of the measurement mapping function of the KF used for reward function learning.
\subsubsection{Adaptation of Measurement Mapping Function via RBFs’ Parameters Update}\label{Sec:RBFs}
As already mentioned, the measurement mapping function needs to be properly set up in KF-based estimations as one of the most important parameters in a Kalman filter. Such a priori knowledge, however, is usually not available, and; consequently it has to be adapted to its correct value. The measurement mapping function employed for the reward weight learning (i.e., $\h$) is modeled by basis functions. Its adaptations, therefore, necessitates recursive update of the basis functions, which is developed below.
As the measurement mapping function depends on a large number of parameters (i.e., $2 \times N$), a gradient descent adaptation approach is used instead of the MMAE model  to adapt these parameters. In this regard, we use the Restricted Gradient Descent (RGD) method~\cite{18},  where partial derivations are exploited to compute the gradient of a defined loss function ($L\k$) in terms of the underlying parameters of the basis functions. We considered the proposed generalized robust lost function in~\cite{Barron} due to its robustness to outliers. Based on Eq.~\eqref{Eq:reward_update}, probability distribution of ($r\k- \bm{\psi}^T(\s\k,a\k)\, \bm{\theta}\k = X\k$) is a Gaussian distribution. The value of shape parameter of the proposed loss~\cite{Barron} is, therefore, considered as $2$, and the loss for RBFs' parameters update will approach to L2 loss function in the limit:
\begin{eqnarray}
 L\k = \left(r\k - \bm{\psi}^T(\s\k,a\k)\, \bm{\theta}\k \right)^2 \label{nEq:35}
\end{eqnarray}
Here, the focus is on minimization of the  loss function ($L\k$) defined in Eq.~\eqref{nEq:35}. To achieve this goal, the required gradients with respect to the mean ($\um$) and covariance ($\Sig$) of RBFs are computed using the chain rule as follows
\begin{eqnarray}
\Delta\um &=& \frac{\partial L\k}{\partial\um} = \frac{\partial L\k}{\partial \bm{\psi}}\frac{\partial \bm{\psi}}{\partial\um}, \\
\text{and}~~\Delta\Sig &=&\frac{\partial L\k}{\partial\Sig} =  \frac{\partial L\k}{\partial \bm{\psi}}\frac{\partial \bm{\psi}}{\partial\Sig} ,
\end{eqnarray}
where the partial derivations are calculated as
\begin{eqnarray}
\frac{\partial L\k}{\partial \bm{\psi}}&=& -2\,{\bm{\theta}\k}^T {\left(L\k\right)}^{\frac{1}{2}}, \\
\frac{\partial \bm{\psi}}{\partial\um}&=&\bm{\psi}\Sig^{-1} (\s\k -\um),\\
\frac{\partial\bm{\psi}}{\partial\Sig} &=& \bm{\psi}\Sig^{-1} (\s\k -\um) (\s\k -\um)^T\Sig^{-1}.
\end{eqnarray}
The partial derivatives are then used to update the means and covariances of  RBFs  as
\begin{eqnarray}
\um\nk^{n} &=& \um\k^{n} -\lambda_{\um}\Delta\um  =  \um\k^{n} +  2\lambda_{\um}{\left(L\k\right)}^{\frac{1}{2}} \bt^T\k \bm{\psi}\big(\Sig\k^{n}\big)^{-1} (\s\k -\um\k^{n}),  \label{Eq:37}\\
\Sig\nk^{n} &=&  \Sig\k^{n} -\lambda_{\Sig}\Delta\Sig = \Sig\k^{n} + 2\lambda_{\Sig}{\left(L\k\right)}^{\frac{1}{2}} \bm{\theta}^T\k  \bm{\psi}\big(\Sig\k^{n}\big)^{-1} (\s\k -\um^{n}\k) (\s\k -\um^{n}\k)^T\big(\Sig\k^{n}\big)^{-1}, \label{Eq:38}
\end{eqnarray}
where $\lambda_{\um}$ and $\lambda_{\Sig}$ shown adaptation rate of $\um$ and $\Sig$, respectively. For stability of the underlying system, according to~\cite{18}, at each step, we only execute one of the updates in Eqs.~\eqref{Eq:37} and~\eqref{Eq:38}. More specifically, we update error covariance matrices of the RBFs via Eq.~\eqref{Eq:38} in scenarios where the covariance is decreasing in size (i.e., $L\k^{\frac{1}{2}} (\bt\k^T \bm{\psi}) < 0$). Otherwise, the means are updated based on Eq.~\eqref{Eq:37}. It is worth mentioning that the above approach avoids unlimited spread of the RBFs error covariance matrices.
\subsection{KTD-based SR Learning}\label{Sec:SR}
As explained in Section~\ref{sec:AKF-SR}, we apply KTD algorithm, which is the combination of TD and Kalamn filter, for the SR learning due to its advantages over DNN-based frameworks.  The TD approach of Eq.~\eqref{Eq:TD_SR2} is, therefore, used to approximate the SR as follows
\begin{eqnarray}
\m^{\pi}_{\text{new}} (\s\k,a\k) \approx \bm{\psi}(\s\k,a\k) +\gamma \m^{\pi}_{\text{old}}(\s\nk,a\nk).\label{nEq:27}
\end{eqnarray}
By reordering terms in Eq.~\eqref{nEq:27}, state-action feature vector $\bm{\phi}(\s\k,a\k)$ can be considered as the measurement model within the KTD framework as follows
\begin{eqnarray}
\bm{\psi}(\s\k,a\k) =\m^{\pi}_{\text{new}}(\s\k, a\k) - \gamma \m^{\pi}_{\text{old}}(\s\nk, a\nk) + \bm{e}\k, \label{Eq:10}
\end{eqnarray}
where $\bm{e}\k$ is considered to be a zero-mean Gaussian noise with the covariance of $\bm{E}\k$. By jointly considering Eqs.~\eqref{Eq:linear_SR} and~\eqref{Eq:10}, the feature vector $\bm{\phi}(\s\k,a\k)$ can be approximated~as
\begin{eqnarray}
\!\!\!\!\!\!\!\!\!\!\!\!\bm{\psi}(\s\k,a\k) &\!\!=\!\!& \W\k\, \bm{\psi}(\s\k,a\k) \!-\! \gamma\, \W\k \bm{\psi}(\s\nk,a\nk)+\bm{e}\k \nonumber \\
\!\!\!\!\!\!\!\!\!\!\!\!&\!\!=\!\!& \W\k \underbrace{\left\{\bm{\psi}(\s\k,a\k)\!-\! \gamma \bm{\psi}(\s\nk,a\nk)\right\}}_{\g\k}+\bm{e}\k. \label{Eq:11}
\end{eqnarray}
To form a KF-based estimate of matrix $\W\k$, we construct a column vector $\w\k$ by stacking the columns of matrix $\W\k$. Based on the vector-trick property of Kronecker product, Eq.~\eqref{Eq:11} can be rewritten as follows
\begin{eqnarray}
\bm{\psi}(\s\k,a\k) &=& (\g^T\k \otimes \I)\w\k + \bm{e}\k, \label{Eq:Kronecker}
\end{eqnarray}
where $\otimes$ denotes the Kronecker product and $\I$ is an identity matrix of a suitable dimension. Eq.~\eqref{Eq:Kronecker} defines the observation of the system \big($\bm{\psi}(\s\k,a\k)$\big) as a linear function of vector $\w\k$, which is to be estimated. Similar to the evolution model of the reward function's parameter $\bt$ in Eq.~\eqref{Eq:reward_predict}, the following  linear dynamic model is defined to represent evolution of  $\w\k$ over time
\begin{eqnarray}
\w\k &=& \A\k \w\pk + \u\k \label{Eq:SR_predict},
\end{eqnarray}
where $\A\k=0.9\I$ and the forcing term, $\u\k$, is considered to be a white Gaussian noise with zero mean and design covariance of $\U\k$. In a similar fashion to that of the reward weight estimation approach of Eqs.~\eqref{Eq:16}-\eqref{Eq:20}, a priori estimates of $\w\k$ and its covariance matrix $\C\k$ at time step $k$ are obtained and then updated using measurement vector $\bm{\psi}(\s\k,a\k)$. Matrix $\W\k$ can be then reconstructed via reshaping the estimated vector $\w\k$ into a $L\times L$ matrix. Now, the state-action value function for each pair ($\s\k,a\k$) can easily be computed using Eq.~\eqref{Eq:Q_estimation}.

\subsection{Active Learning Scheme}
\label{sec:Active Learning}
\begin{table}[t]
\caption{\small Parameters of the proposed $\SR$ framework.}\label{Table:1}
\centering
\begin{tabular}{l |l}
\hline
\textbf{\!\!\!\ Name\!\!\!} & \textbf{Symbol} \\
\hline
Discount factor & $\gamma$ \\
Number of possible actions & $N_{\text{actions}}$ \\
Number of states' variables & $D$ \\
Order of RBFs & $O_{\text{RBF}}$  \\
Number of RBFs & $N_{\text{RBF}}$  \\
RBF's mean & $\um^n$ \\
RBF's covariance & $\Sig^n$ \\
Length of the state-action feature vector & $L$  \\
Number of KFs in MMAE scheme & $N_{\text{KF}}$ \\
Process noise covariance of reward learning  & $\bm{B}$ \\
Measurement noise variance of reward learning  & $\Omega$ \\
Posteriori estimate covariance of reward learning  & $\P$ \\
Process noise covariance of SR learning  & $\U$ \\
Measurement noise covariance of SR learning  & $\bm{E}$ \\
Posteriori estimate covariance of SR learning  & $\C$ \\
Adaptation rate for RBF's mean update& $\lambda_{\um}$ \\
Adaptation rate for RBF's covariance update & $\lambda_{\Sig}$ \\
\end{tabular}
\end{table}
A critical existing challenge in RL problems is the exploration/exploitation trade-off, i.e., selection between exploiting a familiar action for a known reward or exploring unfamiliar actions for unknown rewards. Computation of uncertainty associated with the state-action value function is a key advantage of the proposed $\SR$ learning framework against its DNN-based counterparts. Actions are selected based on computed uncertainty at each filtering step. In other words, at each step, the action leading to the largest decrease in uncertainty of the state-action value function is selected. Since, the state-action value function ($Q(\s\k,a\k)$), has been modeled as a function of estimates $\bt\k$ (reward's weight vector) and $\W\k$ (SR's weight matrix), their uncertainty can be used to approximate uncertainty of the state-action value function. Using the information form of KF (also referred to as information filter~\cite{AK2}), the information associated with $\bt\k$ and $\W\k$, which are denoted by the inverse of their posteriori covariance matrices (i.e., $\P$ and $\C$) are updated as follows
\begin{eqnarray}
\P^{-1}_{k} &=& \P^{-1}_{k|k-1} +\h\k^T \Omega\k^{-1}\h\k, \label{Eq:info:theta}\\
\C^{-1}_{k} &=& \C^{-1}_{k|k-1} +\g\k^T \bm{E}\k^{-1}\,\g\k\label{Eq:info:SR}
\end{eqnarray}
The choice of actions only affects terms ($\h^T\k \h\k$) and ($\g^T\k \g\k$) in Eqs.~\eqref{Eq:info:theta} and~\eqref{Eq:info:SR} as it changes $\h\k$ and $\g\k$. Therefore, $\h\k$ and $\g\k$ can change uncertainty of the state-action value function by choosing different actions. Since the information matrix $\h^T\k \h\k$ cannot be maximized; therefore, its trace (tr($\h^T\k \h\k$) $=$ $\h\k \h^T\k$) will be maximized. As stated previously, $\h\k = \bm{\psi}(\s\k,a\k)$, where vector $\bm{\psi}(\s\k,a\k)$ has been constructed by placing the action-independent vector $\bm{\phi}(\s\k)$ in the corresponding spot for action $a\k$ and setting the feature values for the other actions to zero. The value of $\h\k \h^T\k$ at the specific state $\s\k$, therefore, would be the square norm of the state feature vector $\bm{\phi}(\s\k)$ for different choices of actions (i.e., action-independent). The action $a\k$ is then selected by maximizing the SR weight's information as follows
\begin{eqnarray}
a\k &=& \arg\max_a \Big(\g^T\k (\s\k,a)\g\k(\s\k, a) \Big).\label{Eq:40}
\end{eqnarray}
This contribution is shown in the next section to effectively improve the performance of learning process in terms of achieved rewards from the environment.

The proposed $\SR$ framework is briefed in Algorithm~\ref{algo:1}, and the model's parameters are listed in Table~\ref{Table:1}.
\begin{algorithm}[!t]
\caption{\textproc{The $\SR$ Framework}}
\label{algo:1}
\begin{algorithmic}[1]
\State \textbf{Learning Phase:}
\State \textbf{Input:} $ \gamma, \bm{B}_k, \U_{k}, \lambda_{\um},  \lambda_{\Sig}, \bm{E}\k, \text{and}\, \{\Omega\i_{k} \}$ for $i={1,2,\hdots,N_{\text{KF}}}$
\State \textbf{Initialize:} $\bt_0, \P_{0},\w_0, \C_{0}, {\um_0^{n}, \text{and}\, \Sig_0^{n}}$ for $n={1,2,\hdots,N_{\text{RBF}}}$
\State  \textbf{Repeat} (for each episode):
\State \quad Initialize $\s\k$.
\State \quad \textbf{for} $k=1,2, ...$ \textbf{do}:
\State \quad \quad \textit{\textbf{RBF-based Feature Vector Construction:}} Construct  $\bm{\psi}(\s\k,a)$ through Eqs.~\eqref{Eq:phi} and~\eqref{Eq:si}.
\State \quad \quad \textit{\textbf{Active Learning:}} $a\k=\arg\max\limits_a \Big(\g\k^T (\s\k,a)\g\k(\s\k, a) \Big)$.
\State \quad \quad
 Take action $a\k$ , observe $\s_{k+1}$ and $r\k= R(\s\k,a\k)$.
\State \quad \quad \textit{\textbf{MMAE-based Reward Learning:}} Perform  Eqs.~\eqref{Eq:reward_update}-\eqref{eq:n25} to estimate $\bt\k$.
\State \quad \quad \textit{\textbf{KTD-based SR learning:}} Perform KF on Eqs.~\eqref{Eq:Kronecker}-\eqref{Eq:SR_predict} to update $\w\k$.
\State \quad \quad Reshape $\w\k$ to construct matrix $\W\k$.
\State \quad \quad Calculate feature-based SR vector as  $\m^{\pi}(\s\k,a\k)=\W\k \bm{\psi}(\s\k,a\k)$.
\State \quad \quad Calculate the state-action value function  $Q(\s\k,a\k)=\bt^T\k \m^{\pi}(\s\k,a\k)$.
\State \quad \quad \textit{\textbf{Update RBFs Parameters:}} Perform RGD, Eqs.~\eqref{nEq:35}-\eqref{Eq:38}, to achieve $\Sig\nk^{n}$ and $\um\nk^{n}$ for $1 \leq n \leq N_{\text{RBF}}$.
\State \quad \textbf{end for}
\end{algorithmic}
\end{algorithm}
\section{Experimental Results}  \label{sec:Sim}
The proposed $\SR$ framework is evaluated in this section and compared to the state of the art RL algorithms: Deep Q-Network~\cite{7}, Substochastic Successor Representation (SSR) framework~\cite{count}, and Universal Successor Representations (USR)~\cite{Chen}. In order to demonstrate efficacy of the proposed $\SR$ framework in the learning process as well as its rapid adaptation to the reward changes, the following popular RL benchmarks are considered: (1) \textit{Mountain Car}, (2) \textit{Inverted Pendulum}, and; (3) Lunar Lander. \\The classical DQN described by~\cite{7}, uses a convolutional neural network for calculation of state-action value function. The TD error is then used to perform a gradient descent step with respect to the parameters of the network.
The SSR~\cite{count} computes the norm of the SR while it is being learned with combination of the TD learning and a DNN. It has been shown that SSR implicitly counts state visitation, explaining some light for the uncertainty about that state for exploration/exploitation trade-off. The state-action value function is then learned through a DNN model while it considers the agent's next state uncertainty as an exploration bonus obtained from the SSR. Similar to the proposed $\SR$, SSR estimates the uncertainty of the state (associated value function). But, there are two main differences between $\SR$ and SSR: (i) SSR learns the SR through a DNN model while as previously mentioned, $\SR$ learns the SR via KTD (i.e., incorporation of the TD and KF), and: (ii) SSR considers the norm of the estimated SR as an exploration factor while $\SR$ uses the covariance of the estimated SR for exploration/exploitation trade-off.
The USR~\cite{Chen} uses a deep learning framework to approximate the SR and incorporate it with actor-critic method to learn the SR. There are two key differences to highlight between the proposed $\SR$ and USR approaches. Firstly, USR adopts actor-critic method, which is a TD method that has a separate memory structure to explicitly represent the policy independent of the value function while our proposed $\SR$ framework is a value-based TD method that estimates the policy by estimating the associated value function. The second key difference is that similar to DQN and SSR schemes, USR uses a DNN for the SR learning while our proposed $\SR$ adopts KTD for the SR learning.
\subsection{Choosing $\SR$ Parameters}  \label{sec:parameters}
For adaptation of RBFs' parameters, generally, finding a proper shape parameter of RBFs  is not easy. The centers of RBFs are typically distributed evenly along each dimension of state vector, leading to $O_{\text{RBF}}^D$ centers for $D$ state variables and an order $O_{\text{RBF}}$ given by user; variance of each state variable $(\sigma^2)$ is often initialized to $\frac{2}{O_{\text{RBF}}-1}$. It has been shown that RBFs only generalize locally-changes in one area of the state space and do not affect the entire state space~\cite{RBF_adapt}. The covariance between two different state variables is thus set as zero. Therefore, $\Sig_0 = \sigma^2$ is a scalar if there is one state variable (i.e., $D=1$); otherwise, it is a $D \times D$ diagonal positive definite matrix with the entries of $\frac{2}{O_{\text{RBF}}-1}$.
The number of state variables for both the Inverted Pendulum ($\s\k=[\theta,\dot{\theta}]$) and Mountain Car ($\s\k=[x,\dot{x}]$) environments used in the manuscript are $D=2$. Since we have selected ($O_{\text{RBF}}=3$), $\Sig^n\k$ for all $3^2=9$ RBFs (i.e, $N_{\text{RBF}}=9$) are initialized as $\Sig_0^n = \I_2$, which is a positive definite (invertible) matrix. For the Lunar Lander, $D=6$ and the value of $O_{\text{RBF}}$ is selected to be $2$. Therefore, $\Sig_0^n = 2 \I_6$ for all $64$ RBFs.
\\
As explained in Sub-section~\ref{Sec:RBFs}, the measurement mapping function in Eq.~\eqref{Eq:reward_update} (i.e., $\h$) is constructed as a function of state-action feature vector $\h=\bm{\psi}^T(\s\k,a\k)$, its adaptation, therefore, requires adaptation of the basis functions parameters $\bm{\mu}\k^n, \bm{\Sigma}\k^n$ for ($1 \leq n \leq N_{\text{RBF}}$). In this regard, a gradient descent adaptation approach is adopted to update either of the means and covariances of the RBFs at each time step $k$ based on Eqs.~\eqref{Eq:37} and~\eqref{Eq:38}, respectively. The four terms in the right-hand side of Eq.~\eqref{Eq:38} (i.e., $\big(\Sig\k^{n}\big)^{-1} (\s\k -\um^{n}\k) (\s\k -\um^{n}\k)^T\big(\Sig\k^{n}\big)^{-1}$) form a positive semi-definite matrix and the multiplication of four terms in the left-hand side of Eq.~\eqref{Eq:38} (i.e., $2\lambda_{\Sig}{\left(L\k\right)}^{\frac{1}{2}} \bm{\theta}^T\k  \bm{\psi}$) results in a scalar positive number; therefore, $(-\lambda_{\Sig}\Delta\Sig)$ is a positive semi-definite matrix. Since, the covariance matrix has been initialized to a positive definite matrix ($\frac{2}{O_{\text{RBF}}-1} \I_D$), the updated $\Sig\k^n$ for $k \geq 1$ achieved from Eq.~\eqref{Eq:38} is always a positive definite matrix (consequently, invertible). 

The values for $\lambda_{\bm{\mu}}$ and $\lambda_{\Sig}$ are selected in such a way to keep the system stable.  The discount factor denoted by $\gamma$ affects how much weight is given to the future rewards in the value function. A discount factor $\gamma=0$ will result in state/action values representing the immediate reward, while a higher discount factor will result in the values representing the cumulative discounted future reward that an agent is expected to receive (behaving under a given policy). Commonly (as is the case in the  implemented environments), a large portion of the reward is earned upon reaching the goal. To prioritize this final success, we expect an acceptable $\gamma$ to be close 1. In our manuscript, $\gamma$ is set to the high values of $0.95$ and $0.99$.
 In order to use the KFs implemented for reward learning and SR learning  components of the proposed $\SR$ framework, the parameters of two filters have to be chosen: the variances/covariances of the observation noises, the priors and the variances/covariances of the process noise. The priors ($\bt_0$ and $\w_0$) should be initialized to the values close to the ones looks optimal, or to a default value (i.e., the zero vector). The priors $\P_0$ and $\C_0$ respectively show the certainty in the prior guess of $\bt_0$ and $\w_0$, the lower the more certain. The process noise covariance of a KF is a design parameter. If some knowledge about non-stationarity is available, it can be used to choose this matrix. However, such a knowledge is generally difficult to obtain in advance. The proposed $\SR$ algorithm systematically allows to use a set of different values of process noise covariance ($\bm{B}$ and $\bm{U}$). The priors $\bt_0$, $\w_0$, $\P_0$, and $\C_0$ of the proposed $\SR$ framework are chosen by trial and error. They may not be the best ones, but orders of magnitude are correct. Finally, as stated previously,  the measurement noise covariance of a KF, is one of the most important parameters to be identified. To select this parameter for the KF used in reward function learning process (i.e., $\Omega$), our intuition in the proposed $\SR$ framework is to use MMAE scheme, which covers the potential range of the measurement noise variance $\Omega$ using $N_{\text{KF}}$ mode-matched filters. The parameter $N_{\text{KF}}$ is set a-priori denoting the number of candidates $\Omega^i$, for ($1 \leq i \leq N_{\text{KF}}$). In the experiments, we have selected $N_{\text{KF}}$ to be equal to $11$. It is worth mentioning that measurement noise covariance of the KF used in KTD for the SR learning process ($\bm{E}$), can be also selected via MMAE in a similar way to $\Omega$. But, since $\bm{E}$ is a square high dimensional matrix ($L \times L$), applying MMAE comes with high computational cost. The measurement noise covariance $\bm{E}$ is, therefore, selected by trial and error.
\begin{figure}
\begin{subfigure}{7cm}
\includegraphics[scale=0.6]{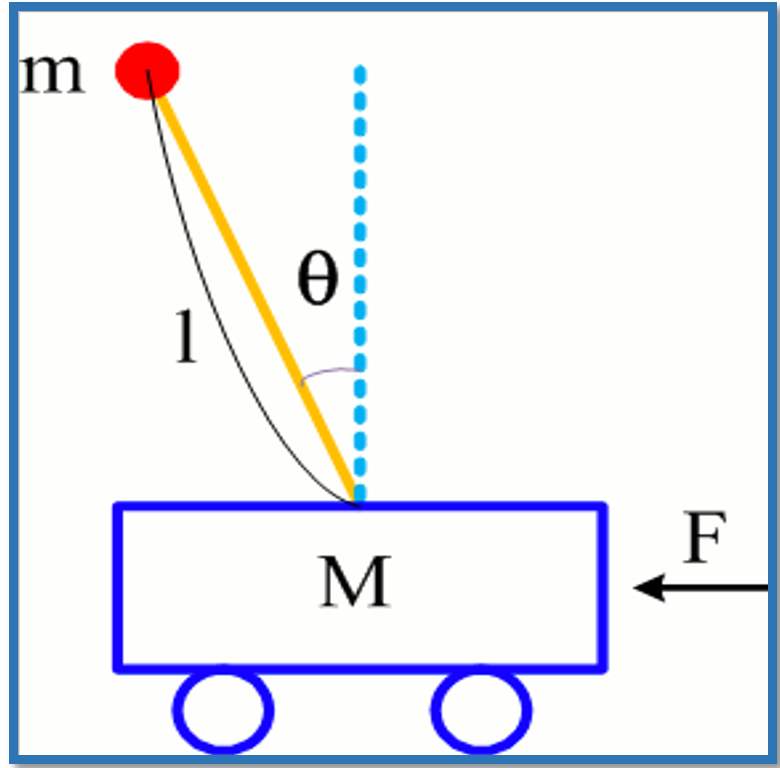}
\caption{}\label{Fig:1}
\end{subfigure}
\quad 
\begin{subfigure}{8cm}
\includegraphics[scale=0.53]{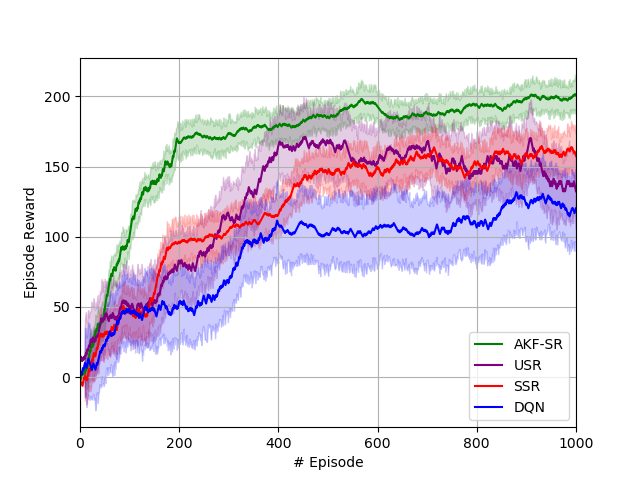}
\caption{ }\label{Fig:reward_Pendulum}
\end{subfigure}
\caption{(a) The Inverted Pendulum environment. (b) The mean (solid lines) and standard deviation (shaded regions) of cumulative episode's reward on the Inverted Pendulum environment over the learning process. }
\end{figure}
\subsection{Inverted Pendulum}  \label{sec:cart_pole}
In the first experiment, Inverted Pendulum environment, shown in Fig.~\ref{Fig:1}, is considered. A pole is attached to a cart moving along a friction-less track. At each time step, an agent can move the cart to the left or to the right. The goal is to prevent it from falling over. The pendulum's angle from upright position and its angular velocity (i.e., $\s\k = [\theta, \dot{\theta}]^T$) constitute the  state of the system. A reward of $+1$ is given at each step that the pendulum is above the horizontal line. Reward $0$ will be provided to the agent when  $|\theta| >\pi/2$, i.e., when the pendulum in the next state is beyond the horizontal. Each episode terminates upon falling of the pendulum or when its length approaches  $200$ runs.

For implementation of the proposed $\SR$, we use $9$ RBFs together with a bias parameter. The size of the state-action feature vector $\bm{\psi}(\s\k,a\k)$ is then equal $30$ as $\mA =\{-50, 0, +50 \}$, i.e., three actions are possible. Initialization is performed as follows
\begin{eqnarray}
\um_0^{n} &\in& \{-\pi/4, 0, +\pi/4 \}\times \{-0.5, 0, +0.5\}\\
\Sig_0^{n} &=& \I_2,
\end{eqnarray}
where $\I_2$ denotes an identity matrix of dimension of $2\times 2$. Based on the Eq.~\eqref{Eq:si}, the state-action feature vector is given by
\begin{eqnarray}
\bm{\psi}(\s\k,a\k=+50)\!=\! [1,\phi_{1},\ldots ,\phi_{9}, 0, \ldots 0,0,\ldots, 0]^T \\
\bm{\psi}(\s\k,a\k=-50)\! = \! [0, \ldots ,0,1,\phi_{1},\ldots \phi_{9},0, \ldots ,0 ]^T\\
\bm{\psi}(\s\k,a\k=0)\!=\![0, \ldots ,0,0, \ldots ,0,1,\phi_{1},\ldots \phi_{9}]^T,
\end{eqnarray}
where the value of $\phi_{n}$, for ($1 \leq n \leq 9$), is computed based on Eq.~\eqref{Eq:phi}. We use the initial values of $\lambda_{\u} = 200$ and $\lambda_{\Sig} =100$ for initialization, which are obtained via trial and error to have a stable system. We use $\gamma = 0.95$ for the discount factor. The process noise covariances for learning of the reward weight and the SR weight are selected as time-invariant values $\bm{B}_{k}=10^{-3}\I_{30}$ and $\U_{k}=10^{-2}\I_{900}$. The measurement noise covariance of SR learning ($\bm{E}_{k}$) is set to $\I_{30}$ and the measurement noise variance candidates for estimation of the reward's weight comes from
\begin{eqnarray}
\Omega\i_{k} \in \{0.01,0.1, 0.2, 0.5, 1, 2, 5, 10, 20, 50, 100\}.
\end{eqnarray}
Finally, $\bt_0 = \bm{0}$ and $\w_0 = \bm{0}$ are used to initialize the weights, and $\P_{0} = 10\I_{30}$ and $\C_{0} = 10\I_{900}$ are utilized to initialize the posteriori covariance matrices.

 A zero-mean Gaussian distribution with standard deviation of $0.1$ is used to randomly select starting angle for each episode. Learning process is performed over $1,000$ episodes. In order to evaluate effect of the proposed uncertainty-based action selection process within the $\SR$ framework, we measure the averaged achieved rewards at each episode over $50$ runs to investigate how the proposed method helps the agent to choose the optimum action, which leads to the least future penalties (maximum future rewards). By selecting such an action at each step, the agent will achieve its goal by doing as few interactions as possible. The achieved reward at each episode, therefore, demonstrates how far the selected actions at each step are from the optimum ones. To illustrate effectiveness of the proposed $\SR$ framework, we compare the results obtained from that with those obtained from the DQN~\cite{7}, SSR~\cite{count}, and USR~\cite{Chen}. As Fig.~\ref{Fig:reward_Pendulum} shows, $\SR$ algorithm performs better than others in terms of achieved rewards toward keeping the inverted pendulum above the horizontal line. These results demonstrate the positive effect of uncertainty usage in action selection process of $\SR$. For a RL algorithm to be usable in different applications, it should be able to reproduce consistent performances across multiple training runs (i.e., reproducibility). As it can be observed in Fig.~\ref{Fig:reward_Pendulum}, at each episode, DNN-based frameworks have a greater variance across 50 runs than $\SR$, which can show sensitivity of DNNs to different factors such as random initialization of the optimization and hyper-parameter setting. Such a high sensitivity can lead to unpredictable performance in practical scenarios and also a large search (consequently, more time and more cost) to find the best model. Another aspect for measuring the reliability of an algorithm is the ability of the algorithm in generating stable performance across episodes (i.e., stability). As can be seen from the results, $\SR$ has less sudden changes in its performance over episodes in comparison to others, which makes $\SR$ framework more reliable to apply to different scenarios. For evaluating the reliability of an RL algorithm, the time and effort spent for tuning of the algorithm's parameters should also be considered. While a DNN has a large number of parameters to be tuned (usually million number of parameters), the proposed filtering-based $\SR$ framework only requires to tune the process and measurement noise covariances ($\bm{B}, \U, \Omega, \bm{E} $), means and covariances of RBFs ($\um^{n}, \Sig^{n}$), and adaptation rates used for RBFs parameters tuning ($\lambda_{\um},  \lambda_{\Sig}$). Much less time and effort is, therefore, required for parameters tuning of $\SR$ framework as compared to its DNN counterparts.
\begin{figure}
  \begin{subfigure}{8cm}
\includegraphics[scale=0.55]{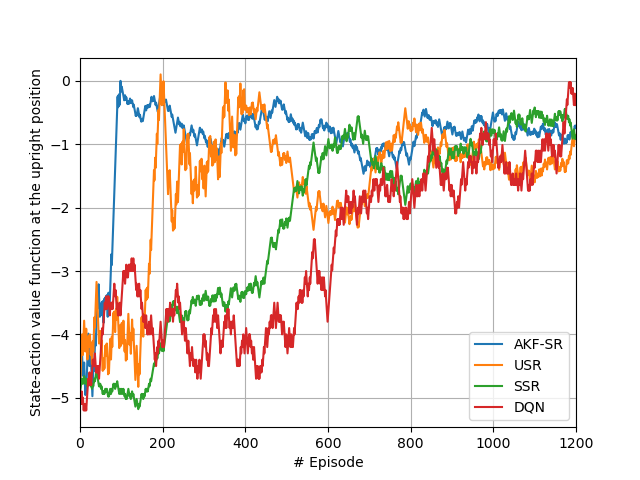}
\caption{}\label{Fig:reward_adaptive}
\end{subfigure}
\quad 
  \begin{subfigure}{8cm}
\includegraphics[scale=0.55]{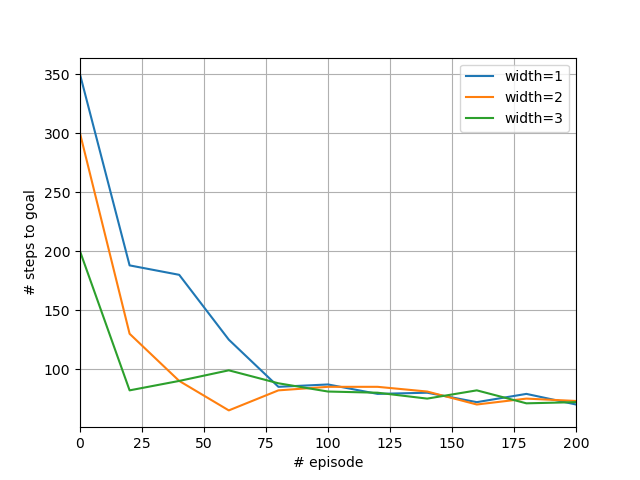}
\caption{ }\label{Fig:stab}
\end{subfigure}
\caption{ (a) State-action value function after change of the reward's value at the upright position of the Inverted Pendulum. (b) Stability analysis of the RBFs.}
\end{figure}

In addition, the key advantage of the proposed SR-based method, which is the decomposition of state-action value function into the SR and reward function, allows the proposed $\SR$ framework to rapidly adapt to changes in the reward function. In order to investigate this strength, after convergence of both models, value of the reward upon keeping the pendulum above the horizontal line is changed from $+1$ to $+3$ and the state-action value is relearned through different frameworks. The estimated state-action value function in the SR context of $\SR$ and USR~\cite{Chen}, will converge to this change by just updating (relearning) the reward given the new external rewards (the SR remains same and the agent does not need to learn the SR another time). However, adjustment of DQN and SSR algorithm to changes comes with updating cached values at all states in the environment. Results shown in Fig.~\ref{Fig:reward_adaptive} confirm that the computing the value function via a dot product between the SR and reward enables the agent to adapt itself rapidly to the new value function by just updating the reward function.

Furthermore, to illustrate the stability of the RBFs in the manuscript, in the Inverted Pendulum task, we have fixed the number of RBFs to be $10$. We have performed proposed $\SR$ scheme for $200$ episodes. The entire process have been repeated $50$ times for three different values of widths $(\Sig)$ of RBFs and the number of steps to goal was averaged over $50$ times. As Fig.~\ref{Fig:stab} shows, by using the RBFs for the reward function and the SR approximations, we can achieve a steady state performance.
\subsection{Mountain Car}  \label{sec:car}
\begin{figure}[t!]
  \begin{subfigure}{7cm}
\includegraphics[scale=0.55]{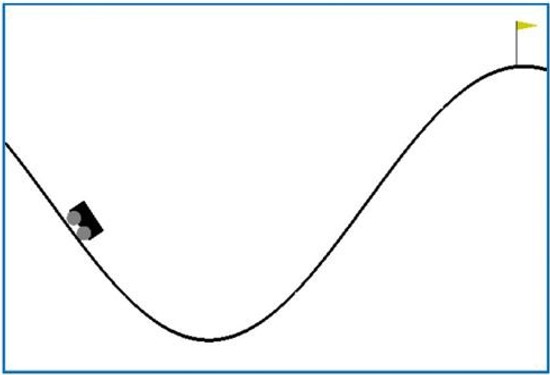}
    \caption{}
    \label{Fig:4}
  \end{subfigure}
  \quad
  \begin{subfigure}{9cm}
\includegraphics[scale=0.6]{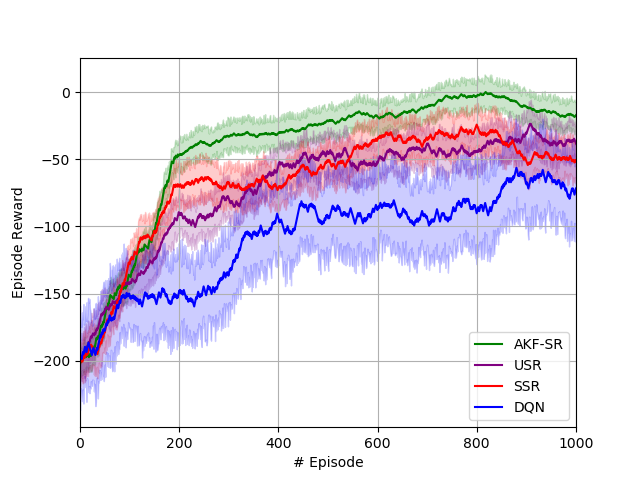}
    \caption{}
    \label{Fig:Car_reward}
  \end{subfigure}
\caption{ (a) The Mountain Car environment.  (b) The mean (solid lines) and standard deviation (shaded regions) of cumulative episode's reward on the Mountain Car environment over the learning process.}\end{figure}

The Mountain Car benchmark RL platform shown in Fig.~\ref{Fig:4} is used to conduct the  second experiment. In this classic RL problem, one car is positioned on a 1-D track between two mountains. The objective of the RL model is to bring the car to the top of the right mountain. It is assumed that the car cannot climb the right mountain in one single try. For success in this scenario, therefore, the car needs to go back and forth building up momentum to reach its goal. Position and velocity of the car (i.e., $\s=[x, \dot{x}]^T$) constitute the state of the system. Available actions are ``push left'', ``push right'', and ``no push'', represented by set $\mA =\{0, 1,2 \}$. The objective of the RL model is to reach the top position on the right mountain, i.e., car's position becomes greater than or equal to $0.5m$. In scenarios that the objective is not achieved (i.e., car can not mount the right hill), a $-1$ reward will be given to the agent for each step. No penalty will be assigned in case that the car reaches the left hill (this state is treated as hitting a wall). The starting state of each episode is a random position within $-0.6m$ to $-0.4m$ range with zero velocity.

In a similar fashion to the Inverted Pendulum experiment in Sub-section~\ref{sec:cart_pole}, the state-action feature vector for Mountain Car is also constructed based on $9$ RBFs and a bias parameter (resulting in a feature vector of size $30$). Initialization is performed as follows
\begin{eqnarray}
\um_0^{(n)} &\!\!\!\! \in\!\! \!\! & \{-0.775, -0.35, +0.775 \}\times \{-0.035, 0, +0.035\}\,\,\,\,\,\,\,\,\,\,\,\,\,\,
\\
\Sig_0^{(n)} & \!\!\!\! = \!\!\!\!& \I_2
\end{eqnarray}
The process noise covariances for learning the reward's weight and the SR weight are selected as $\bm{B}\k=10^{-2}\I_{30}$ and $\U\k=10^{-2}\I_{900}$. The discount factor $\gamma$ is set to $0.95$. Values $\bt_0 = \bm{0}$ and $\w_0 = \bm{0}$ are used to initialize the weights. Values $\P_{0} = 10\I_{30}$ and $\C_{0} = 10\I_{900}$ are exploited to initialize the posteriori covariance matrices. The measurement noise covariance of SR learning ($\bm{E}_{k}$) and the measurement noise variance candidates for estimation of the reward's weight are selected from the same set as the Inverted Pendulum environment.  The model learns from $1,000$ episodes, with a maximum steps of $200$ at each episode. Each episode is repeated $50$ times and averaged. Fig.~\ref{Fig:Car_reward} demonstrates that $\SR$ outperforms DNN-based frameworks in the terms of performance criteria defined for Inverted Pendulum environment, i.e., cumulative reward received at each episode, time and computational cost, and reproducibility and stability aspects of the algorithms.
\subsection{Lunar Lander}  \label{sec:lunar_lander}
\begin{figure}[t!]
  \begin{subfigure}{8cm}
\includegraphics[scale=0.45]{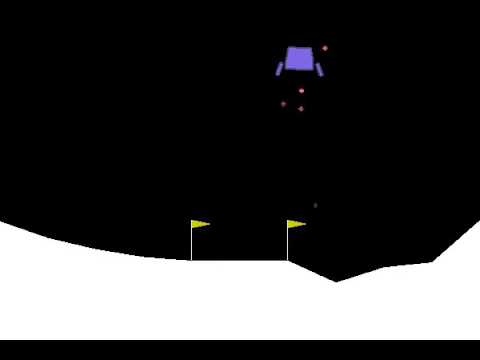}
\caption{}\label{Fig:Lunar}
  \end{subfigure}
  \quad
  \begin{subfigure}{8cm}
\includegraphics[scale=0.55]{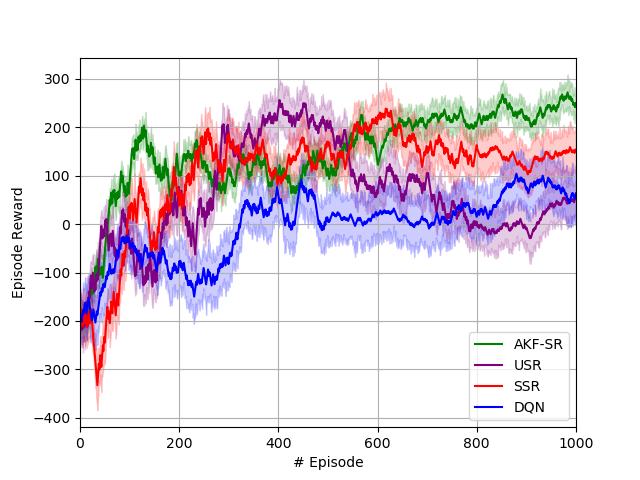}
\caption{ }\label{Fig:Lunar_reward}
  \end{subfigure}
  \caption{(a) The Lunar Lander environment. (b)  The mean (solid lines) and standard deviation (shaded regions) of cumulative episode's reward on the Lunar Lander environment over the learning process.}
\end{figure}

\begin{figure}[t!]
  \begin{subfigure}{8cm}
    \centering
    \includegraphics[width=7.5cm]{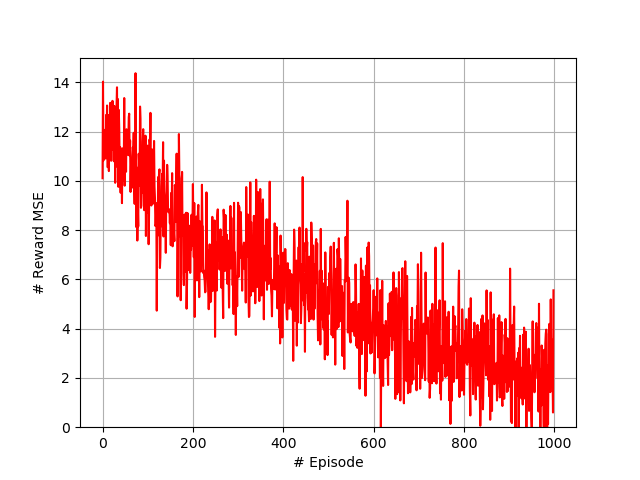}
    \caption{}
  \end{subfigure}
  \quad
  \begin{subfigure}{8cm}
    \centering
    \includegraphics[width=7.5cm]{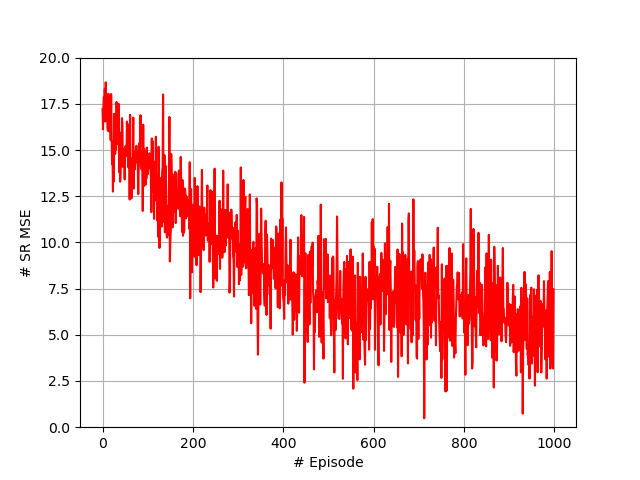}
    \caption{}
  \end{subfigure}
  \caption{\small The calculated MSE loss of the estimated reward function and the SR in the $\SR$ framework over 1000 episodes for the Lunar Lander: (a) MSE of the estimated reward, and (b) MSE of the estimated SR.}\label{Fig:loss}
\end{figure}

In the third experiment, we focus on the Lunar Lander environment, which is a more complicated environment compared to the two previous ones. In the  Lunar Lander environment, the goal for the RL agent is to learn to land successfully on a landing pad located at coordinate  $(0,0)$ in a randomly generated surface on the moon as shown in Fig.~\ref{Fig:Lunar}. The state space of the system consists of the agent's position $(x,y)$ in space, horizontal and vertical velocity $(v_x,v_y)$, orientation in space $\theta$, and angular velocity $\dot{\theta}$. The agent has four possible actions, i.e., do nothing; firing the left engine, firing the main engine, and; firing the right engine ($\mA =\{0, 1, 2, 3 \}$).  Reward for landing on the pad is about $100$ to $140$ points, varying on the lander placement on the pad. If the lander moves away from the landing pad it loses reward. Each episode terminates if the lander lands or crashes, receiving additional +$100$ or -$100$ points, respectively. Each leg ground contact worth +$10$ points. Firing the main engine results in a -$0.3$ point penalty for each frame. The problem is considered solved if the agent receives +$200$ points over $100$ iterations. The RBFs of order two are considered for each state variable resulting in $64$ RBFs for each action. Consequently, the size of the feature vector $\bm{\phi}(\s\k,a\k)$ will be $256$. Based on the useful range of each variable of state vector, the initial mean and covariance of the RBFs are chosen as follows
\begin{eqnarray}
\u_0^n &\in& \{-0.333, +0.333\}^6,
\\
\Sig_0^n&=& 2\I_{6}.
\end{eqnarray}
The initial values of $\lambda_{\bm{mu}}$ and $\lambda_{\Sig}$ are both selected as $200$ to keep the system stable. The discount factor is selected as $0.99$. The  noise covariances are selected as $\bm{B}\k=10^{-2}\I_{256}$ and $\bm{E}\k=\I_{256}$. Candidate values for $\Omega$ are selected from the same set as was used for the Inverted Pendulum environment. Like the two previous environments, the agent is trained through different $1000$ episodes and training process is repeated $50$ runs. Fig.~\ref{Fig:Lunar_reward} depicts the cumulative reward averaged over $50$ runs reward per each episode for different approaches. Due to the complexity of the lunar lander environment, the cumulative rewards achieved from $\SR$ has more fluctuations in comparison to inverted pendulum and mountain car. But as it can be observed, $\SR$ still outperforms its counterparts in terms of cumulative reward received at each episode, and reproducibility and stability of the algorithms.
\\
In order to evaluate the linear assumptions of the reward function and the SR, the Mean Squared Error (MSE) loss of the estimated reward and the SR for $1000$ episodes over $50$ training runs are calculated and are shown in Fig.~\ref{Fig:loss}. As expected, both the reward and the SR losses decrease over training process. Despite the fluctuations in the loss values, the linear model estimates seem to be good enough to achieve a better performance over model-free DQN method and SR-based SSR and USR frameworks. The loss of state-action value function, which is summation of the SR loss and the reward loss, are averaged over $1000$ episodes and shown in Table~\ref{Table:2} for different frameworks. As it can be seen in Table~\ref{Table:2}, the average loss in the proposed $\SR$ is less than that of the DNN-based approaches.
\begin{table}[t]
\caption{\small The MSE loss of state-action value function averaged over $1000$ episodes.}\label{Table:2}
\centering
\begin{tabular}{|p{3cm}|p{1.5cm}|p{.7cm}|p{.7cm}|p{.7cm}|}
\hline
\multirow{2}{*}{Environment} & \multirow{2}{*}{$\SR$} & \multirow{2}{*}{DQN} & \multirow{2}{*}{SSR} & \multirow{2}{*}{USR} \\
&&&&\\
\hline
Inverted Pendulum &   $6.43$ & $11.98$ & $7.36$ & $7.69$
\\ \hline
Mountain Car & $6.78$ & $14.78$ & $9.94$ & $8.46$
 \\ \hline
Lunar Lander & $8.93$ & $14.59$ & $16.76$ & $11.98$
\\ \hline
\end{tabular}
\end{table}
\subsection{Algorithmic Complexity}  \label{sec:complexity}
As explained in the previous sections, the proposed $\SR$ framework consists of four main modules: (i) RBF-based Feature Vector Construction, which maps each state $\s\k$ to $N_{\text{RBF}}$-dimensional state feature vector $\bm\phi(\s\k)$ consisting of basis functions achieved from Eq.~\eqref{Eq:phi}. The state-action feature vector $\bm{\psi}(\s\k,a\k)$ is then generated by assigning $\bm{\phi}(\s\k)$ to the corresponding spot for action $a\k$ while the feature vector values for the remainder of the actions are set to zero. The whole construction process of state-action feature vector $\bm\psi(\s\k,a\k)$ has a computational complexity of $O(N_{\text{RBF}}\times D^3)$ per iteration, (ii) Reward Learning, which first, estimates the reward weight vector $\bt\k$ via implementation of $N_{\text{RBF}}$ parallel KFs. The global computational complexity (per iteration) of MMAE is $O(N_{\text{KF}}\times L^3)$. The measurement mapping function of the reward weight vector is then adapted by updating the means and covariances of basis functions through Eqs.~\eqref{Eq:37} and~~\eqref{Eq:38}. Since at each time step, only the mean or covariance is updated, this process results in memory complexity of $O(\frac{L+D^3}{2})$, (iii) KTD-based SR Learning, which estimates the SR weight matrix $\W\k$ via KTD algorithm. This estimation procedure can be computed in $O(L^4)$~\cite{23} per iteration, and; (iv) Active Learning Scheme, which selects the action leading to the largest
decrease in uncertainty of the state-action value function in $O(L \times N_{\text{actions}})$. The global computational complexity (per iteration) of the proposed $\SR$ scheme is, therefore, in $O(N_{\text{RBF}}\times D^3+ N_{\text{KF}}\times L^3 +L^4)$. It should be noted that the main focus of this paper is on the improvement of the performance of RL agents by considering their uncertainties/beliefs for choosing different actions. As it has been shown in the manuscript, this uncertainty can be achieved by using KF for the reward function learning and KTD for the SR learning. The proposed MMAE and RGD improve the performances of the used KF by adaptation of its two most important parameters: measurement noise covariance and measurement mapping function. The computational complexities of DQN~\cite{7}, USR~\cite{Chen} and SSR~\cite{count} are not specified; however, based on our experiments and since the proposed $\SR$ has much less trainable parameters than the DQN~\cite{7}, USR~\cite{Chen}, and SSR~\cite{count} frameworks, which all use DNNs with large number of parameters for the learning process, $\SR$ has less computational cost compared to its DNN-based counterparts. As an example, DQN with only one hidden layer of size $64$ has $11140$ parameters resulting in high computational complexity and memory requirement. Training with a mini-batch of $32$ on a typical performance GPU shows that it needs over $3.5$ GB of local DRAM. 
\section{Conclusion}  \label{sec:con}
This paper proposed a novel KTD-based SR framework, referred to as the $\SR$, to learn goal reaching behavior in RL problems. The $\SR$ learns the value function by computing the inner product of the SR and reward function's weight vector. This factorization of the value function, enables the learned SR to be reused for other tasks with the same domains but different reward values. Moreover, we extended KTD framework of MF algorithms~\cite{23} to the SR learning, which estimates uncertainty of the value function, and also deals with over-fitting, parameter sensitivity (consequently, unreliability in reproducing consistent performances across multiple runs), and high memory requirement problems of DNN-based algorithms. Another key advantage of the proposed $\SR$ algorithm is using RGD and an innovative MMAE schemes for the reward function learning, which tackles the effect of improper selection of the filter's parameters on its performance and stability to make the framework more usable in practical problems. Furthermore, an innovative active learning scheme, which considers uncertainty of the value function achieved from KTD algorithm for actions selection, was adopted within the $\SR$ framework to deal with the exploration/exploitation dilemma. The proposed $\SR$ framework was evaluated based on three continuous state space RL benchmarks: Inverted Pendulum, Mountain Car, and Lunar Lander. The averaged accumulative reward and MSE loss for $1000$ episode over 50 runs were calculated. The results showed that the proposed $\SR$ framework outperforms its counterparts in terms of cumulative reward, reproducibility and stability aspects of reliability, and time and effort spent for finding and learning the best model. To compare adaptation speed of $\SR$ with DNN-based frameworks to changes in the reward function, the reward value of Inverted Pendulum environment was changed and all algorithms were relearned. As it was expected, $\SR$ could adapt itself to the changes faster than other frameworks.

For estimating of the SR weight matrix $\W\k$ through implemented Kalman filter within the KTD framework, we mapped the matrix into a column vector by stacking
its columns one underneath the other and proceed with standard Kalman filter for vectors. However, it is often difficult to ensure that this vectorized estimation method does not cause any consequential losses in the original structure of the problem. Furthermore, for high-dimensional estimation models (e.g., a RL environment with a large number of states variables), such a method comes with an excessive computational cost. A number of studies have investigated to develop estimation algorithms in terms of the original system matrices~\cite{Matrix-KF,Nissen}. As future work, we plan to apply matrix-based filtering algorithms for estimation of the SR weight matrix in order to extend the proposed $\SR$ framework to more complex RL tasks with high-dimensional states in an efficient manner.
\\
\bibliographystyle{model1-num-names}

\end{document}